\documentclass[10pt,twocolumn,letterpaper]{article}
\usepackage{iccv}
\usepackage{my}

\iccvfinalcopy 

\ificcvfinal\fi

\begin{document}

\title{Swin3D: A Pretrained Transformer Backbone \\for 3D Indoor Scene Understanding}

\author{
{Yu-Qi Yang}$^{1}$\thanks{Interns at Microsoft Research Asia. \textsuperscript{\dag}Contact person.}  \qquad
{Yu-Xiao Guo}$^{2}$\qquad
{Jian-Yu Xiong}$^1$\textsuperscript{*} \qquad
{Yang Liu}$^{2\dag}$ \\
{Hao Pan}$^{2}$\qquad
{Peng-Shuai Wang}$^{3}$\qquad
{Xin Tong}$^{2}$\qquad
{Baining Guo}$^{2}$ \\
\small {$^1${Tsinghua University}\quad \quad
  $^2${Microsoft Research Asia}  \quad\quad
  $^3${Peking University}}  \\
{\tt\small \{t-yuqyan, yuxgu, v-jixiong, yangliu\}@microsoft.com} \\
{\tt\small haopan@microsoft.com\quad wangps@hotmail.com\quad
\{xtong,bainguo\}@microsoft.com}
}

\maketitle
\ificcvfinal\thispagestyle{empty}\fi

\begin{abstract}
  The use of pretrained backbones with fine-tuning has been successful for 2D vision and natural language processing tasks, showing advantages over task-specific networks. In this work, we introduce a pretrained 3D backbone, called {\SST}, for 3D indoor scene understanding. We design a 3D Swin transformer as our backbone network, which enables efficient self-attention on sparse voxels with linear memory complexity, making the backbone scalable to large models and datasets. We also introduce a generalized contextual relative positional embedding scheme to capture various irregularities of point signals for improved network performance. We pretrained a large {\SST} model on a synthetic Structured3D dataset, which is an order of magnitude larger than the ScanNet dataset. Our model pretrained on the synthetic dataset not only generalizes well to downstream segmentation and detection on real 3D point datasets, but also outperforms state-of-the-art methods on downstream tasks with +2.3 mIoU and +2.2 mIoU on S3DIS Area5 and 6-fold semantic segmentation, +1.8 mIoU on ScanNet segmentation (val), +1.9 mAP@0.5 on ScanNet detection, and +8.1 mAP@0.5 on S3DIS detection. A series of extensive ablation studies further validate the scalability, generality, and superior performance enabled by our approach. 
\end{abstract}

\section{Introduction}
A paradigm shift has been seen in the fields of Natural Language Processing (NLP) and 2D vision, where the use of large pre-trained backbones has been highly successful~\cite{dosovitskiy2020image,bao2022beit,devlin2018bert,brown2020language,liu2021swin,liu2022swin}. This approach has the advantage of being able to generalize to various tasks, while also reducing the amount of network design and training needed, as well as the amount of labeled data required for various vision tasks. This involves pre-training a backbone network with a simple design on a large dataset, and then fine-tuning it for different downstream tasks. However, the development of generic and scalable pretrained 3D backbones for point cloud understanding is still in its early stages, and existing pretrained 3D backbones are not as effective as the state-of-the-art non-pretrained approaches for many 3D vision tasks.

This paper seeks to investigate the scalability and generality of a 3D pretrained model, without the need for a complex network design. We introduce a pretrained 3D backbone, {\SST}, for 3D indoor scene understanding tasks. Our method uses sparse voxels to represent the 3D point cloud of an input 3D scene and adapts the network design of Swin Transformer \cite{liu2021swin}, which was originally designed for regular 2D images, to unorganized 3D points as the 3D backbone. We identify two key issues that prevent the na\"{i}ve 3D extension of Swin Transformer from exploring large models and achieving high performance: \textit{the high memory complexity} and \textit{the ignorance of signal irregularity}. To address these issues, we develop a novel 3D self-attention operator to compute the self-attentions of sparse voxels within each local window. This reduces the memory cost of self-attention from quadratic to linear with respect to the number of sparse voxels within a window and computes efficiently without sacrificing self-attention accuracy. Additionally, it enhances self-attention by capturing various signal irregularities by generalizing \emph{contextual relative positional embedding}~\cite{wu2021rethinking,lai2022stratified} to point signals.

Our novel {\SST} backbone design allows us to scale up the backbone model and utilize the large amount of data for pretraining. To demonstrate this, we pretrained a large {\SST} model with \SI{60}{M} parameters on a 3D semantic segmentation task using a large synthetic 3D dataset: Structured3D~\cite{zheng2020structured3d}. This dataset is ten times larger than the ScanNet dataset~\cite{dai2017scannet} that contains \SI{21}{K} rooms. After pretraining, we combined the pretrained {\SST} backbone with task-specific back-end decoders and fine-tuned the models for various 3D indoor scene understanding tasks.

    We evaluated the performance of our method on both 3D detection and semantic segmentation tasks on real data including ScanNet and S3DIS datasets. Experimental results show that our {\SST} pretrained on the synthetic dataset exhibits good generality and outperforms all existing state-of-the-art methods with +8.1 mAP@0.5 on S3DIS detection, +2.2 mIoU on 6-fold S3DIS segmentation~\cite{S3DIS}, +1.9 mAP@0.5 on ScanNet detection and +1.8 mIoU on ScanNet segmentation (validation set) respectively. We carefully analyzed the contributions of different factors (\eg model size, data size, pretraining) to the performance of our model. Our studies show that our pretrained backbones with fine-tuning are superior to the same models trained from scratch and outperform other existing models pretrained with the same large data significantly.

The large backbone model and the large amount of 3D data used for pretraining enabled by our backbone design are critical to the superior performance of our method in the downstream tasks. We believe that our work illustrates the great potential of a unified pretrained backbone to various 3D understanding tasks. To facilitate and inspire future research along this path, we have released our code and trained models at \url{https://github.com/microsoft/Swin3D}.

\section{Related Work}

\myparagraph{Vision transformers} Transformers based on the attention mechanism have been used successfully in computer vision and have achieved great results in many 2D vision tasks, such as image classification, semantic segmentation, and object detection (\cf the comprehensive surveys~\cite{khan2021transformers,han2022survey,guo2022attention}). The plain vision transformer~\cite{dosovitskiy2020image} computes global self-attention over the entire image, thus providing long-range attention between image patches, but this leads to high memory and computational costs due to the quadratic complexity of self-attention. To address this issue, local-window self-attention over small non-overlapping patches~\cite{han2021demystifying,liu2021swin} was introduced. Various techniques have been proposed to improve the long-range attention of window-based self-attention, such as using window hierarchy~\cite{liu2021swin}, constructing non-local self-attention patterns~\cite{dong2022cswin,tu2022maxvit,zhang2022vsa,wu2022pale,yang2021focal,li2022sepvit,wang2021crossformer,chu2021twins}, and expanding the receptive field via convolution~\cite{chen2022mixformer,yuan2021hrformer}. Most vision transformers are pretrained with large-scale image datasets and serve as generic vision backbones for multiple purposes.

\myparagraph{3D transformers for point cloud understanding} Transformers have been quickly adapted to 3D~\cite{lahoud20223d}. Guo \etal~\cite{guo2021pct} used global attention on points and achieved good results in object classification and shape segmentation. Zhao \etal~\cite{zhao2021point} introduced local attention on point clouds, which reduced memory and computational complexity and enabled the point transformer to be used for point clouds at the scene level. Wu \etal~\cite{wu2022point} further improved the point transformer by using grouped vector attention and partition-based pooling. Fast Point Transformer~\cite{park2022fast} employed voxel hashing and a lightweight self-attention layer to enhance network efficiency. Stratified Transformer~\cite{lai2022stratified} adapted the Swin Transformer design for 3D point clouds and proposed the stratified strategy to expand the reception field and used contextual relative positional encoding~\cite{wu2021rethinking} to strengthen self-attention with position information. In 3D vision, most 3D transformers that have been created have been tailored to particular tasks, and simply training them on a large amount of data does not result in better performance, as we evaluated (\cref{sec:results}).

\myparagraph{Pretrained 3D backbones} Self-supervised learning strategies have been applied to 3D backbone pretraining. PointContrast~\cite{Xie2020} used point-level losses to pretrain a sparse-convolution-based 3D U-Net. MID-Net~\cite{Wang2020a} pretrained an Octree-based HRNet with multiresolution contrastive losses. Hou \etal~\cite{hou2020efficient} improved the efficacy of PointContrast by taking advantage of spatial information. DepthContrast~\cite{zhang2021self} employed depth maps to enhance contrastive learning. Masked-signal-modeling-based transformer models, such as BEiT~\cite{bao2022beit} and the masked autoencoder~\cite{HeCXLDG22}, have been used for 3D pretraining~\cite{Liu2022maskdis,zhang2022point,Yu_2022_CVPR,Pang2022MaskedAF}.  Wu Wu \etal ~\shortcite{wu2023masked} intergrated the masked point modeling strrategy with contrastive learning to boost backbone pretraining. Recently, pretrained image or CLIP models have been utilized for 3D learning~\cite{wang2022p2p,dong2022autoencoders,huang2022clip2point,huang2022frozen}, forming a new type of pretrained 3D backbones. ShapeNet~\cite{shapenet2015} and ScannNet~\cite{dai2017scannet} are the main 3D data sources for the above pre-training work. Despite the rapid development of 3D pretraining, its performance on 3D indoor scene understanding is still inferior to that of the state-of-the-art non-pretrained approaches.

\section{Architecture overview}

\subsection{Naive 3D extension of Swin Transformer}
Hierarchical window-based transformers, such as Swin Transformer~\cite{liu2021swin}, are widely used in generic vision due to their high efficiency, multiscale feature learning, scalability, and improved performance compared to 2D CNN backbones. It is thus a logical step to extend Swin Transformer-like architectures for 3D point cloud learning. The implementation of window attention mechanism in 3D appears to be straightforward --- partition the input 3D point cloud into non-overlapping 3D windows and compute self-attention on nonempty-voxel features within regular and shifted windows. However, as Lai~\etal \shortcite{lai2022stratified} observed in their ablation study, this naive extension does not lead to superior performance. We have identified two key issues that explain this uninspiring performance: \emph{memory complexity} and \emph{signal irregularity}.

\myparagraph{Memory complexity} The quadratic complexity of self-attention leads to a high memory cost in 3D. For a 3D window with size $M\times M \times M$, the average number of non-empty voxels within the window is approximately $\mathcal{O}(M^2)$, thus the memory cost of executing vanilla self-attention within the window is about $\mathcal{O}(M^4)$. For a 3D scene, the number of windows $N_w$ could be considerable, depending on the size of the 3D scenes. Therefore, the total memory cost $\mathcal{O}(N_wM^4)$ in 3D could be much higher than its 2D counterpart, where the image size is usually low. This memory issue prevents the use of large windows and more Swin layers, making it difficult to design large models that can benefit from large data. \looseness=-1

\begin{figure}[t]
    \centering
    \begin{overpic}[width=1\linewidth]{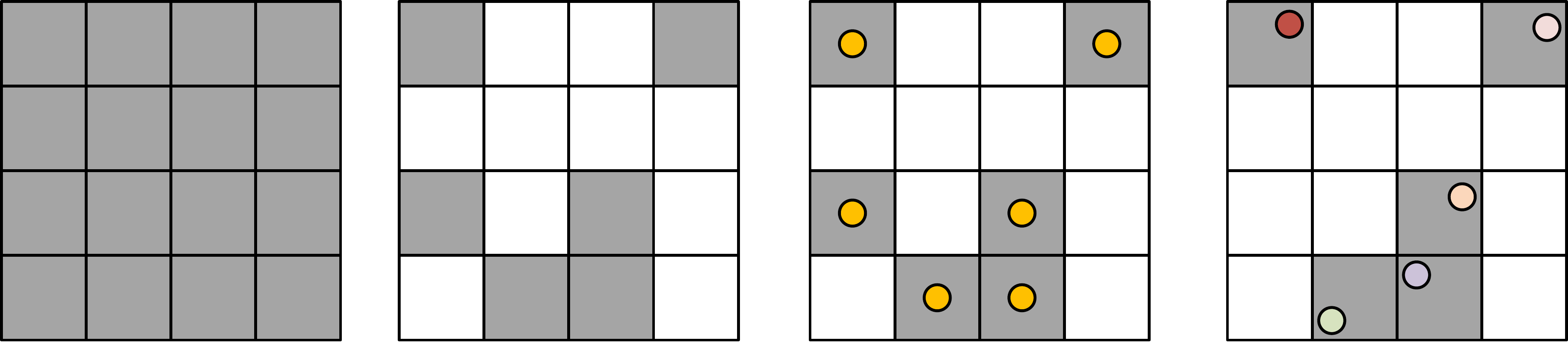}
        \put(8.8,-3.5){\small \textbf{(a)}}
        \put(33.8,-3.5){\small \textbf{(b)}}
        \put(60,-3.5){\small \textbf{(c)}}
        \put(85.7,-3.5){\small \textbf{(d)}}
    \end{overpic}
    \vspace{1px}
    \caption{A 2D illustration of sparse points in a $4\times 4$ window. \textbf{(a)} A fully-occupied window. \textbf{(b)} A sparsely-occuppied window, with white cells being empty. \textbf{(c)} Regularly-distributed sparse points in a window. \textbf{(d)} The sparse points are irregularly distributed in the window, and different circle colors indicate the varying point-wise signal, such as the RGB color. For simplicity, only one point is drawn on non-empty cells. }
    \label{fig:sparse} 
\end{figure}

\myparagraph{Signal irregularity} The locations of 3D points can be highly irregular: points can be found anywhere within their occupied voxel, while for 2D visual transformers, image pixels are regularly distributed at grid cell centers. Additionally, since points are usually equipped with other raw signals, such as RGB colors, if we consider both point positions and other pointwise signals as voxel signals, the signal irregularity, \ie, the relative signal variation between any two voxels in a window, can be quite varied. Previous works~\shortcite{lai2022stratified} address point irregularity only by using positional encoding in self-attention, but they are unaware of the variations of other signals. In \cref{fig:sparse}, we illustrate sparse voxels in a window and signal irregularity on a 2D example.

\begin{figure*}[t]
    \centering
    \begin{overpic}[width=1\linewidth]{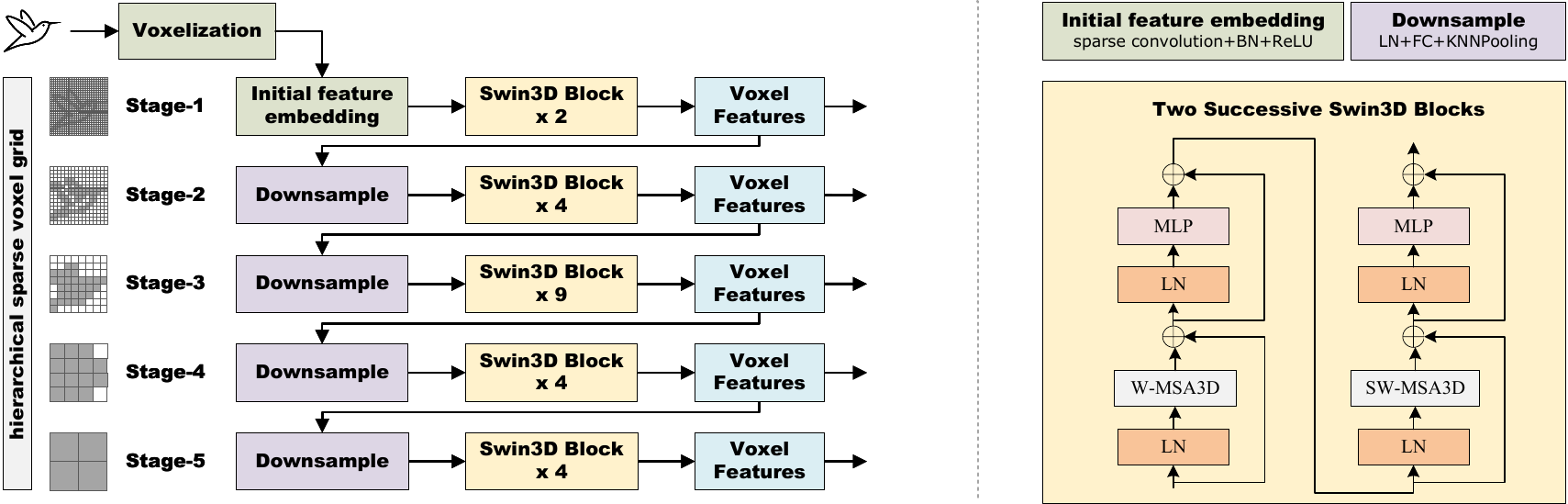}
        \put(54, 23){\scriptsize $N_1\times C_1$}
        \put(54, 17.5){\scriptsize $N_2\times C_2$}
        \put(54, 12){\scriptsize $N_3\times C_3$}
        \put(54, 6){\scriptsize $N_4\times C_4$}
        \put(54, 0.6){\scriptsize $N_5\times C_5$}
    \end{overpic}
    \caption{ \textbf{Left}: The architecture of {\SST}. It consists of five-stage transformer blocks that apply self-attention to sparse voxels within regular and shifted windows at various levels of a hierarchical sparse grid. The grids on the left are the illustration of sparse grids in 2D, with gray cells representing non-empty voxels. $N_i$ denotes the number of sparse voxels at the $i$-th level, and $C_i$ is the feature channel dimension. \textbf{Right}: The detailed operations of each module.}
    \label{fig:sswin} 
\end{figure*}
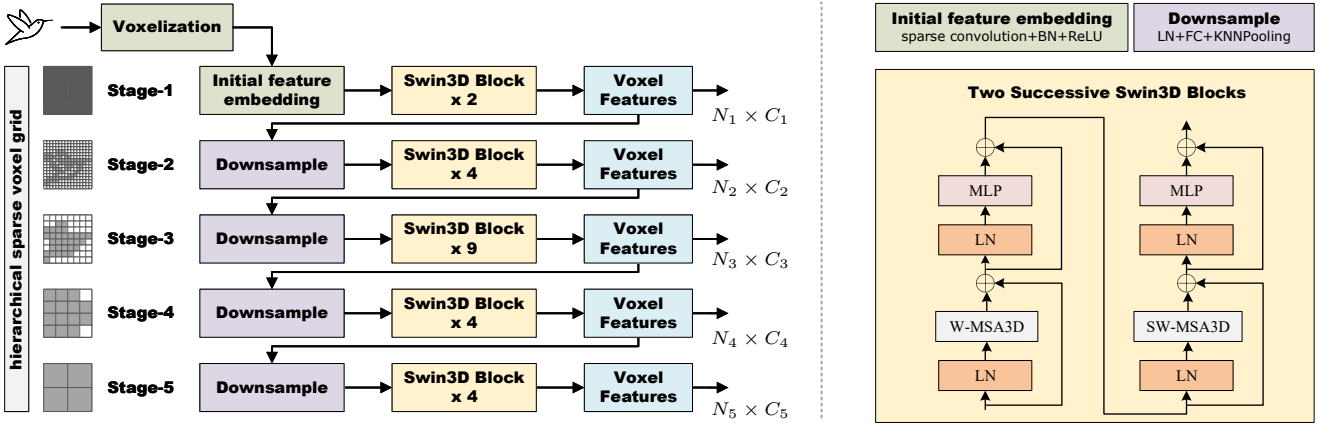

\subsection{{\SST} Architecture}
We have designed our {\SST} backbone to address the issues mentioned above. It has a hierarchical structure similar to the Swin Transformer and is composed of the following modules: \emph{voxelization}, \emph{initial feature embedding}, \emph{{\SST} block}, and \emph{downsample}. The \emph{voxelization} module discretizes an input point cloud into a multiscale sparse voxel grid, the \emph{initial feature embedding} module generates sparse voxel features at the finest voxel level for feature attention, the \emph{{\SST} block} performs memory-efficient self-attention on sparse voxel features within regular and shifted windows and addresses signal irregularity by \emph{contextual relative signal encoding}, and the \emph{downsample} module aggregates the sparse voxel features at $l$-th level to $(l+1)$-th level. {\SST} contains 5-stage {\SST} blocks, each of which operates at different voxel resolution. It serves as a multiscale feature encoder for any input point cloud and can be easily combined with task-specific decoders for various 3D tasks. \cref{fig:sswin} illustrates the architecture of our backbone. The details of each module are presented in the following section.

\section{Module design of {\SST} Transformer} \label{sec:module}

\subsection{Voxelization} \label{subsec:voxelization}

\myparagraph{Point cloud input} An input point cloud is typically associated with point-wise signals, such as point position, point color and point normal. For a point $\bp$, the concatenation of these signals is represented by $\bs_{\bp} \in \mathbb{R}^m$. For any color component or normal component, it is mapped to the range $[-1,1]$.  A common setting is $m=6$, meaning 3D point coordinates and RGB color.

\myparagraph{Voxelization} We employ sparse voxels as point proxies in our backbone. A 5-level hierarchical sparse voxel grid is constructed from the input point cloud, as shown in \cref{fig:sswin} with a 2D example. By default, the voxel size at the finest level is set to \SI{2}{\centi\meter} for indoor scenes. The voxel size is doubled when the voxel level is increased by one. Point information is stored in the voxels from the finest level to the coarsest level in the following manner.
\begin{enumerate}[leftmargin=*]
    \item[-] For the voxel $\bv$ at the finest level, we randomly select one point from within $\bv$ and designate it as the \emph{representative point} of $\bv$, which is denoted by $\br_v$.
    \item[-] For the voxel $\bv$ at the $(l+1)$-th level, we first select all representative points from its child voxels. We then choose the representative point closest to the center of these points and make it the representative point of $\bv$.
\end{enumerate}
The voxelization step assigns unstructured points to structured sparse voxel grids, and the representative point is used to supply raw features to the initial feature embedding and provide contextual information for computing {\SST} self-attention (see \cref{subsec:block}). For simplicity, we denote the signal at the representative point of voxel $\bv$ as $\bs_{\bv}$.

\subsection{Initial feature embedding} \label{subsec:initembedding}
Motivated by the observation that using a linear layer or MLP to project raw features to a high dimension does not yield good performance for Swin-like transformer architectures \cite{lai2022stratified}, we propose to lift the raw feature via sparse convolution. At the finest voxel level, we apply one layer of sparse convolution with a $3\times3\times 3$ kernel, followed by a batch normalization (BN) and a ReLU layer, to transform the input voxel feature to $\mathbb{R}^{C_1}$. The input feature on voxel $\bv$ is set as the concatenation of the positional offset: $\br_v - \bc_v$ and other point signals stored at $\br_v$, where $\bc_v$ is the center of $\bv$. We do not use the absolute point position because our goal is to learn local priors via convolution. Compared to KPConv~\cite{Thomas2019} utilized by \cite{lai2022stratified}, our initial feature embedding is much lighter and five times faster.

\subsection{{\SST} block} \label{subsec:block}
Our {\SST} block is based on the Swin Transformer block design: it operates on both regular and shifted windows in 3D. The voxel grid is split into non-overlapping windows at level $l$, with a window size of $M \times M \times M$. The shifted window is created by shifting the regular window with an offset of $\bigl(\lfloor \frac{M}{2}\rfloor,\lfloor \frac{M}{2}\rfloor,\lfloor \frac{M}{2}\rfloor\bigr)$. We modified the standard multi-head self-attention to improve memory efficiency and address signal irregularity, as described below. The number of heads is denoted by $N_H$.

\subsubsection{Memory-efficient self-attention scheme}

\myparagraph{Vanilla self-attention}
For a given input window with $N$ non-empty voxels, denoted by $\{\bv_i\}_{i=1}^N$, the network features of these voxels are represented by $\{\mf_i\}_{i=1}^N$. A vanilla multi-head self-attention is then applied to the input voxel features, which is a weighted sum of the projected voxel features:
\begin{equation}\label{eq:attention}
    \mf^\star_{i,h}  = \sum_{j=1}^N \alpha_{ij,h} \cdot \mf_j \bV, \qquad i=1,\ldots, N,
\end{equation}
 where $\{\mf^\star_{i,h}\}_{i=1}^N$ are the output feature vectors at the $h$-th head. The weight coefficient $\alpha_{ij,h}$ is the SoftMax version of $e_{ij,h}$, \ie $\exp(e_{ij,h})/\sum_{k=1}^N \exp(e_{ik,h})$, and $e_{ij,h}$ is in a scaled dot-product attention form:
 \begin{equation}
    e_{ij,h} = \dfrac{(\mf_i \bQ) (\mf_j \bK)^T}{\sqrt{d}}.
\end{equation}
Here, $\bQ, \bK, \bV$ are linear projection matrices for \emph{Query}, \emph{Key} and \emph{Value} computation, and $d$ is the channel number of the $h$-th head.

\myparagraph{Memory-efficient self-attention}
The common multi-head self-attention implementation requires two passes to calculate $\{\alpha_{ij,h}\}$. The first pass computes all $\exp(e_{ij})$s and the second pass accumulates their sums, \ie $\sum_{k=1}^N\exp(e_{ik})$. This necessitates storing all $\alpha_{ij}$s for \cref{eq:attention}, resulting in $\mathcal{O}(N^2 \times N_H)$ memory complexity. For Swin-like Transformer architectures, this complexity is $\mathcal{O}(N^2 \times N_H \times N_w)$, where $N_w$ is the number of windows. In the case of a 3D Swin Transformer with window size $M\times M \times M$, $N = \mathcal{O}(M^2)$ and $N_w$ can be very large if the scale of the input point cloud varies greatly. This high memory cost in 3D limits the use of large windows and deeper networks; however, this is not a significant issue in the 2D Swin Transformer because $N_w$ is at least one order smaller than its 3D counterpart, and the input image usually has a fixed size.

We observe that \cref{eq:attention} can be rewritten in the following form:
\begin{equation}\label{eq:attention-modify}
\mf^\star_{i,h} = \dfrac{\sum_{j=1}^N \bigl(\exp(e_{ij,h}) \mf_j \bV\bigr)}{\sum_{j=1}^N \exp(e_{ij,h})}, \qquad i=1,\ldots, N,
\end{equation}
which allows us to postpone the SoftMax normalization and avoid constructing and storing $\{\alpha_{ij,h}\}$ explicitly. Therefore, we modify the second pass by calculating the denominator and numerator of \cref{eq:attention-modify} simultaneously. During computation, $\{\exp(e_{ij,h})\}_{j=1}^N$ for $\mf^\star_{i,h}$ are calculated on the fly, without storage. This eliminates the quadratic complexity of the memory cost of $\{\alpha_{ij,h}\}$. For gradient propagation, each $\exp(e_{ij,h})$ is computed twice during the training stage. However, this additional computation cost is negligible as the self-attention computation is a memory-intensive operation not a computation-intensive operation, thus our memory reduction does not slow down the computation and could reduce the execution latency as well (see evaluation in \cref{sec:results}).

\subsubsection{Contextual relative signal encoding}

Swin Transformer utilizes \emph{relative position bias}~\cite{shaw2018self} to enhance the performance of its backbone. Wu \etal~\cite{wu2021rethinking} proposed a novel contextual mode for relative positional encoding, referred to as \emph{contextual relative position encoding} (cRPE), which adds relative position encoding to both queries and keys. Lai \etal~\cite{lai2022stratified} used cRPE to capture fine-grained position information in 3D self-attention computation. In our work, we extended cRPE to all kinds of signals, not just point positions, as other signals such as RGB color also display high variation within a window and these variations should be captured by self-attention. We refer to this generalized version as \emph{contextual relative signal encoding}, or cRSE. The multi-head self-attention with cRSE is formulated as follows.

We first modify $e_{ij}$ to include the difference between the voxel signals $\Delta \ms_{ij} := \bs_{\bv_i}-\bs_{\bv_j}$:
\begin{equation}
e_{ij,h} = \dfrac{(\mf_i \bQ) (\mf_j \bK)^T + b_{ij,h}}{\sqrt{d}}.
\end{equation}
Here, $b_{ij,h}$ is the contextual signal encoding:
\begin{align}
   b_{ij,h} = &(\mf_i \bQ) \bigl(\mt_{K,h}(\Delta \ms_{ij})\bigr)^T + \nonumber \\& (\mf_j \bK) \bigl(\mt_{Q,h}(\Delta \ms_{ij})\bigr)^T.
\end{align}

The output of self-attention is revised to:
\begin{equation}
\mf^\star_{i,h} = \dfrac{\sum_{j=1}^N \exp(e_{ij,h}) (\mf_j \bV +  \mt_{V,h}\bigl(\Delta \ms_{ij})\bigr)}{\sum_{j=1}^N \exp(e_{ij,h})},
\end{equation}
where $\mt_{K,h}$, $\mt_{Q,h}$ and $\mt_{V,h}$ are trainable functions that map the signal differences to $\mathbb{R}^d$.

In order to make these trainable functions lightweight, we follow \cite{shaw2018self, lai2022stratified} to quantify signal differences by a set of learnable look-up tables: $\{t^{Q,h}_1, \ldots, t^{Q,h}_m\}$, $\{t^{K,h}_1, \ldots, t^{K,h}_m\}$, and $\{t^{V,h}_1, \ldots, t^{V,h}_m\}$, each with a fixed length $L_i$, where $\star$ can be $Q$, $K$ or $V$. For an input vector $\Delta$, its table indices are determined by: \begin{equation} I_l(\Delta) = \bigl\lfloor \frac{( \Delta[l] - \texttt{minquat}[l] )L_l}{\texttt{quat}[l]} \bigr\rfloor. \end{equation} $\Delta[l]$ is the $l$-th component of $\Delta$, $\texttt{quat}[l]$ and $\texttt{minquat}[l]$ are the quantification range and the lower bound of signal difference for the $l$-th signal, respectively. For common signal types, $\texttt{quat}[l]$ and $\texttt{minquat}[l]$ are defined as follows:
\begin{enumerate}[leftmargin=*]
\item[-]  If the $l$-th signal corresponds to point position, $\texttt{quat}[l] = 2 h$ and $\texttt{minquat}[l] = -h$, where $h$ is the physical height of the cubic window.
\item[-] If the $l$-th signal corresponds to one of the RGB components, $\texttt{quat}[l] = 2$ and $\texttt{minquat}[l] = -1$.
\item[-] If the $l$-th signal corresponds to one of the point normal components, $\texttt{quat}[l] = 2$ and $\texttt{minquat}[l] = -1$.
\end{enumerate}
With the look-up tables and index functions, we have $\mt_{Q,h} (\Delta) = \sum_{l=1}^m t^Q_{l,h}[I_l(\Delta)]$, $\mt_{K,h} (\Delta) = \sum_{l=1}^m t^Q_{l,h}[I_l(\Delta)]$, $\mt_{V,h} (\Delta) = \sum_{l=1}^m t^V_{l,h}[I_l(\Delta)]$. The use of look-up tables for cRSE introduces additional $3\sum_{i=1}^m L_i \times N_H $ parameters. By default, we set $L_l = 4$ if the $l$-th signal corresponds to point position, and $L_l = 16$ if the $l$-th signal corresponds to color or normal components. The effectiveness of cRSE is evaluated extensively in \cref{subsec:ablation}.

\subsubsection{Transformer block} Our {\SST} transformer block, denoted by \texttt{S-MSA3D} (on regular windows) and \texttt{SW-MSA3D} (on shifted windows), is composed of the revised multi-head self-attention along with other transformer components such as LayerNorm and MLP layer, as illustrated in \cref{fig:sswin}-right.

\subsubsection{Efficient implementation} \label{subsec:efficent}
We revised the self-attention module of Stratified Transformer~\cite{lai2022stratified} to improve its efficiency. The revision includes optimizing kernel scheduling, enabling half-precision, and reducing accesses of atomic operations, and we call this revision our vanilla implementation. We further extend cRPE to cRSE and integrate our memory-efficient design.  More details on the implementation are presented in Appendix.

\subsection{Downsample} \label{subsec:downsample}
The voxel features at the $l$-th level are downsampled to the $(l+1)$-th level by first lifting all sparse voxel features to $\mathbb{R}^{C_{l+1}}$ through a LayerNorm and an FC layer. Then, for any sparse voxel $\bv$ at the $(l+1)$-th level, the features of its $k$-nearest voxels at the $l$-th level are maxpooled and assigned to $\bv$. This downsampling strategy is referred to as \texttt{KNNPooling}, with $k$ set to $16$ by default.

\section{{\SST} Backbone pretraining}

\subsection{Backbone models} \label{subsec:backbonemodel}
We created two versions of {\SST}: {\SST}-S and {\SST}-L. The window size for the first stage is set to $5\times 5 \times 5$ and for the rest stages, it is $7\times 7 \times 7$. The layer numbers are $\{2, 4, 9, 4, 4\}$ with downsample strides $\{3, 2, 2, 2\}$. The feature dimensions (\#FD) and head numbers (\#HD) at each stage are:
\begin{enumerate}[leftmargin=*]
    \item[-]\textbf{{\SST}-S}:  \#FD = $\{48,96,192,384,384\}$, \#HD = $\{6, 6, 12, 24, 24\}$;
    \item[-]\textbf{{\SST}-L}:  \#FD = $\{80,160,320,640,640\}$, \#HD = $\{10, 10, 20, 40, 40\}$.
\end{enumerate}
By default, the input point signal contains positional and color information only. When the input signal contains point normals and cRSE uses normal signals, we use {\SST}$_n$-S and {\SST}$_n$-L to denote our backbone models.
The efficiency and support for large models of our backbone design is determined by the following assessments.

\paragraph{Model efficiency}
We evaluated the memory and computation efficiency of our backbone model by computing the average memory usage and computational time based on 70 point clouds from ScanNet~\cite{dai2017scannet} (see \cref{tab:ablation-efficiency}). Our {\SST}-S model is referred to as \emph{Our-Efficient}. We also compared it with (1) Stratified Transformer~\cite{lai2022stratified}, which also adapted the Swin Transformer for 3D point clouds; (2) an improved version of Stratified Transformer based on our revision described in \cref{subsec:efficent} and contextual relative signal encoding, excluding our memory-efficient self-attention scheme, referred to as \emph{Our-Vanilla} implementation. All models had the same number of Transformer blocks and the same amount of Transformer parameters. Compared with \cite{lai2022stratified}, our vanilla implementation already significantly decreased computational and memory cost and our memory-efficient self-attention further reduced half of memory consumption and sped up the computation. The memory and computation efficiency of our design allowed us to explore large {\SST} models to take advantage of large data sets.

\begin{table}[t]
    \centering
    \resizebox{\columnwidth}{!}{
        \begin{tabular}{l|r|r|r|r|r|r|r}
            \toprule
                          &               & \multicolumn{2}{c|}{\mythead{Impl. of ~\cite{lai2022stratified}}} & \multicolumn{2}{c|}{\mythead{Our-Vanilla}} & \multicolumn{2}{c}{\mythead{Our-Efficient}}                                                  \\
            \midrule
            \mythead{Block} & \mythead{\#Pts} & \mythead{Time}                                                    & \mythead{Mem.}                             & \mythead{Time}                              & \mythead{Mem.} & \mythead{Time}  & \mythead{Mem.}    \\
            \midrule
            Stage-1       & 109.48\,k     & 487.7                                                           & 1380.1                                   & 25.7                                      & 555.4        & \textbf{20.3} & \textbf{268.68} \\
            Stage-2       & 15.05\,k      & 122.9                                                           & 467.4                                    & 16.0                                      & 180.4        & \textbf{14.1} & \textbf{95.4}   \\
            Stage-3       & 4.01\,k       & 56.5                                                            & 233.6                                    & 9.3                                       & 104.8        & \textbf{7.1}  & \textbf{47.9}   \\
            Stage-4       & 1.01\,k       & 29.0                                                            & 120.5                                    & 6.5                                       & 60.0         & \textbf{4.8}  & \textbf{24.4}   \\
            Stage-5       & 0.25\,k       & 9.6                                                             & 36.7                                     & 4.3                                       & 21.2         & \textbf{2.4} & \textbf{8.7}    \\
            \bottomrule
        \end{tabular}
    }
    \vspace{2pt}
    \caption{We compare the efficiency of self-attention by reporting the average statistics of point number (same to the number of non-empty voxels), execution time (in milliseconds) and memory footprint (in megabytes) for a single forward-backward iteration. The implementation of \cite{lai2022stratified} only works with double-precision.
    }    \label{tab:ablation-efficiency} 
\end{table}

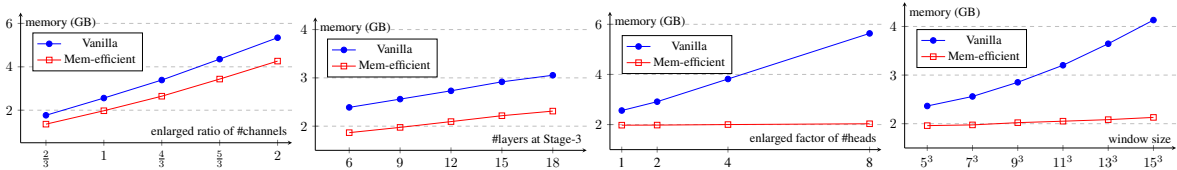
\begin{figure*}[t]
    \centering
    \resizebox{0.9\textwidth}{!}{
    \begin{tikzpicture}[scale=1]
        \begin{axis}[
                unit vector ratio=2 12 1,
                xlabel={\small enlarged ratio of \#channels},
                ylabel={\small memory (GB)},
                axis line style={->},
                axis lines=middle,
                xmin=25, xmax=100,
                ymin=0.5, ymax=6.5,
                xtick=data,
                xticklabels={$\frac{2}{3}$,$1$,$\frac{4}{3}$,$\frac{5}{3}$, $2$},
                ytick={2, 4, 6},
                legend style={at={(0.45,0.55)},anchor=south east},
                ymajorgrids=true,
                grid style=dashed,
            ]
            \addplot[color=blue,mark=*,]coordinates {(32,1.766)(48,2.560)(64,3.390)(80,4.353)(96,5.341)};
            \addplot[color=red,mark=square,]coordinates {(32,1.357)(48,1.975)(64,2.645)(80,3.435)(96,4.263)};
            \legend{\small{Vanilla},\small{Mem-efficient}}
        \end{axis}
    \end{tikzpicture}
    \begin{tikzpicture}[scale=1]
        \begin{axis}[
                unit vector ratio=2 5.7 1,
                xlabel={\small \#layers at Stage-3},
                ylabel={\small memory (GB)},
                axis line style={->},
                axis lines=middle,
                xmin=4, xmax=20,
                ymin=1.5, ymax=4.25,
                xtick={6, 9, 12, 15, 18},
                ytick={2, 3, 4},
                legend style={at={(0.45,0.55)},anchor=south east},
                ymajorgrids=true,
                grid style=dashed,
            ]
            \addplot[color=blue,mark=*,]coordinates {(6,2.389)(9,2.560)(12,2.732)(15,2.918)(18,3.054)};
            \addplot[color=red,mark=square,]coordinates {(6,1.866)(9,1.975)(12,2.095)(15,2.217)(18,2.312)};
            \legend{\small{Vanilla},\small{Mem-efficient}}
        \end{axis}
    \end{tikzpicture}
    \begin{tikzpicture}[scale=1]
        \begin{axis}[
                unit vector ratio=2 8.5 1,
                xlabel={\small enlarged factor of \#heads},
                ylabel={\small  memory (GB)},
                axis line style={->},
                axis lines=middle,
                xmin=4, xmax=50,
                ymin=1, ymax=6.5,
                xtick=data,
                xticklabels={1,2,4,8},
                ytick={2, 4, 6},
                legend style={at={(0.45,0.55)},anchor=south east},
                ymajorgrids=true,
                grid style=dashed,
            ]
            \addplot[color=blue,mark=*,]coordinates {(6,2.560)(12,2.913)(24,3.818)(48,5.637)};
            \addplot[color=red,mark=square,]coordinates {(6,1.975)(12,1.979)(24,2.000)(48,2.032)};
            \legend{\small{Vanilla},\small{Mem-efficient}}
        \end{axis}
    \end{tikzpicture}
    \begin{tikzpicture}[scale=1]
        \begin{axis}[
                unit vector ratio=2 4.3 1,
                xlabel={\small window size},
                ylabel={\small memory (GB)},
                axis line style={->},
                axis lines=middle,
                xmin=4, xmax=16,
                ymin=1.5, ymax=4.5,
                xtick=data,
                xticklabels={$5^3$, $7^3$, $9^3$, $11^3$, $13^3$, $15^3$},
                ytick={2, 3, 4},
                legend style={at={(0.45,0.55)},anchor=south east},
                ymajorgrids=true,
                grid style=dashed,
            ]
            \addplot[color=blue,mark=*,]coordinates {(5,2.364)(7,2.560)(9,2.851)(11,3.202)(13,3.645)(15,4.132)};
            \addplot[color=red,mark=square,]coordinates {(5,1.960)(7,1.975)(9,2.021)(11,2.051)(13,2.083)(15,2.129)};
            \legend{\small{Vanilla},\small{Mem-efficient}}
        \end{axis}
    \end{tikzpicture}
    }
    \caption{Evaluation on the support to large models. \textbf{Top-left}: GPU memory consumption of wider networks. \textbf{Top-right}: GPU memory consumption of deeper networks. \textbf{Bottom-left}: GPU memory consumption with respect to head numbers. \textbf{Bottom-right}: GPU memory consumption with respect to window size.  The GPU memory was measured on a forward-and-backward iteration, averaged on all ScanNet data.
    }
    \label{fig:scalability} 
\end{figure*}

\paragraph{Support to large models} We assessed the scalability of our design by examining its GPU memory consumption when applied to large models. We used \SST-S as the base model and tested its GPU memory utilization on ScanNet data, increasing the width and depth of the network, the number of heads, and the window size. Additionally, we compared the results to the vanilla version of our model, which does not employ memory-efficient self-attention.  The experiment setup is as follows:
\begin{enumerate}[leftmargin=*]
    \item[-]\textbf{Wider networks}:  We created five models based on \SST-S by increasing the feature channels with five different ratios: $\frac{2}{3}, 1, \frac{4}{3}, \frac{5}{3}, 2$, respectively. The model with ratio $\frac{5}{3}$ is equivalent to \SST-L.
    \item[-]\textbf{Deeper networks}: We created five models based on \SST-S by changing the number of Swin block layers at Stage-3 to 6, 9, 12, 15, 18, respectively. Here, the default \SST-S uses 9 layers.
    \item[-]\textbf{Large head number} We created five models based on \SST-S by increasing the number of heads by four different factors: 1, 2, 4, 8.
    \item[-]\textbf{Large window size} We tested different window sizes: $5^3, 7^3, 9^3, 11^3, 13^3, 15^3$.
\end{enumerate}
We present the GPU memory consumption of the test in \cref{fig:scalability}. We can observe that as the channel number and the block number are proportional to memory storage, the rate of memory usage is almost constant; and the models with our memory-efficient self-attention save around $25\%$ memory compared to the models with the vanilla-version self-attention. The quadratic memory saving by our memory-efficient self-attention is clearly visible when increasing the head number or the window size, compared to the models with the vanilla-version self-attention. In conclusion, the reduction of GPU memory enabled by our memory-efficient self-attention makes our backbone architecture suitable for designing large backbones.

From the test, it is evident that our backbone design can accommodate models with greater capacity than \SST-L. However, we found experimentally that more large models tend to overfit our pre-training data set, so we did not explore them further in this study.

\subsection{Backbone pretraining}\label{subsec:backbonepretrain}

\begin{table*}[t]
    \centering
    \resizebox{0.95\linewidth}{!}{
        \scriptsize
        \setlength\tabcolsep{2.5pt}
        \begin{tabular}{c|c|r|cccccccccccccccccccccccccccccccccccc}
            \toprule
            \mythead{Datasets}         & \mythead{Task}            & \mythead{\#C}         &
            \rotatebox{90}{\footnotesize wall}     &
            \rotatebox{90}{\footnotesize floor}    & \rotatebox{90}{\footnotesize cabinet} & \rotatebox{90}{\footnotesize bed} & \rotatebox{90}{\footnotesize chair} & \rotatebox{90}{\footnotesize sofa} & \rotatebox{90}{\footnotesize table} & \rotatebox{90}{\footnotesize door} & \rotatebox{90}{\footnotesize window} & \rotatebox{90}{\footnotesize bookshelf} & \rotatebox{90}{\footnotesize bookcase} & \rotatebox{90}{\footnotesize picture} & \rotatebox{90}{\footnotesize counter} & \rotatebox{90}{\footnotesize desk} & \rotatebox{90}{\footnotesize shelves} & \rotatebox{90}{\footnotesize curtain} & \rotatebox{90}{\footnotesize dresser} & \rotatebox{90}{\footnotesize pillow} & \rotatebox{90}{\footnotesize mirror} & \rotatebox{90}{\footnotesize ceiling} & \rotatebox{90}{\footnotesize refrigerator} & \rotatebox{90}{\footnotesize television} & \rotatebox{90}{\footnotesize shower curtain} & \rotatebox{90}{\footnotesize nightstand} & \rotatebox{90}{\footnotesize toilet} & \rotatebox{90}{\footnotesize sink} & \rotatebox{90}{\footnotesize lamp} & \rotatebox{90}{\footnotesize bathtub} & \rotatebox{90}{\footnotesize garbagebin} & \rotatebox{90}{\footnotesize board} & \rotatebox{90}{\footnotesize beam} & \rotatebox{90}{\footnotesize column} & \rotatebox{90}{\footnotesize clutter} & \rotatebox{90}{\footnotesize otherstructure} & \rotatebox{90}{\footnotesize otherfurniture} & \rotatebox{90}{\footnotesize otherprop.}                                                     \\
            \midrule
            Structured3D             & Seg.                    & 25                  & \checkmark            & \checkmark           & \checkmark            & \checkmark           & \checkmark             & \checkmark                & \checkmark               & \checkmark              & \checkmark              &                      &                         & \checkmark              &                         & \checkmark             & \checkmark             & \checkmark              & \checkmark                   & \checkmark                 & \checkmark                     & \checkmark                 & \checkmark             & \checkmark           &                      & \checkmark              &                            & \checkmark            & \checkmark           &                        &                         &                                &                                &                           &            & \checkmark & \checkmark & \checkmark \\
            \midrule
            \multirow{2}{*}{ScanNet} & Seg.                    & 20                  & \checkmark            & \checkmark           & \checkmark            & \checkmark           & \checkmark             & \checkmark                & \checkmark               & \checkmark              & \checkmark              & \checkmark           &                         & \checkmark              & \checkmark              & \checkmark             &                        & \checkmark              &                              &                            &                                &                            & \checkmark             &                      & \checkmark           &                         & \checkmark                 & \checkmark            &                      & \checkmark             &                         &                                &                                &                           &            &            & \checkmark &            \\
                                     & Det.                    & 18                  &                       &                      & \checkmark            & \checkmark           & \checkmark             & \checkmark                & \checkmark               & \checkmark              & \checkmark              &                      & \checkmark              & \checkmark              & \checkmark              & \checkmark             &                        & \checkmark              &                              &                            &                                &                            & \checkmark             &                      & \checkmark           &                         & \checkmark                 & \checkmark            &                      & \checkmark             & \checkmark              &                                &                                &                           &            &            &            &            \\
            \midrule
            \multirow{2}{*}{S3DIS}   & Seg.                    & 13                  & \checkmark            & \checkmark           &                       &                      & \checkmark             & \checkmark                & \checkmark               & \checkmark              & \checkmark              &                      & \checkmark              &                         &                         &                        &                        &                         &                              &                            &                                & \checkmark                 &                        &                      &                      &                         &                            &                       &                      &                        &                         & \checkmark                     & \checkmark                     & \checkmark                & \checkmark &            &            &            \\
                                     & Det.                    & 5                   &                       &                      &                       &                      & \checkmark             & \checkmark                & \checkmark               &                         &                         &                      & \checkmark              &                         &                         &                        &                        &                         &                              &                            &                                &                            &                        &                      &                      &                         &                            &                       &                      &                        &                         & \checkmark                     &                                &                           &            &            &            &            \\
            \bottomrule
        \end{tabular}
    }
    \vspace{2pt}
    \caption{We provide a list of semantic labels for our Structured3D pretraining dataset, as well as the datasets of the downstream tasks. The list is denoted by ``\#C'', which stands for the number of segmentation labels. Additionally, ``Seg.'' and ``Det.'' indicate the tasks of semantic segmentation and 3D detection, respectively.}  \label{tab:tasks-category} 
\end{table*}

\myparagraph{Pretraining data preparation} We opt to pretrain our backbones with the Structured3D dataset~\cite{zheng2020structured3d}, which contains 21835 rooms in 3500 \textbf{synthetic} scenes and is equipped with a variety of high-quality 3D objects and layouts. This dataset is one-order larger than other \textbf{real} indoor datasets such as MatterportLayout~\cite{zou2021manhattan} (2295 rooms) and ScanNet~\cite{dai2017scannet} (1613 rooms). We use the provided RGBD and panoramic images to generate the training data as follows.

We acquire all RGBD and panoramic images associated with the room using the official room description, and project RGB and semantic labels of all images into 3D points based on their intrinsic and extrinsic camera parameters. To reduce the large number of points, we divide the space into cubes of \SI{1}{cm^3} and keep only one point in each cube. The remaining points form the point cloud of the room. We also estimate point normals from depth maps for training \SST$_n$ models.  \looseness=-1

\myparagraph{Backbone pretraining} We selected 3D semantic segmentation as our pre-training task, with $25$ semantic labels. Four original segmentation labels (counter, box, toilet, bathtub) were excluded due to their rarity. \cref{tab:tasks-category} shows the segmentation labels used in pretraining, as well as those in the downstream tasks of ScanNet segmentation and detection, and S3DIS segmentation and detection. There is some overlap between our pre-training data and the datasets of the downstream tasks, and some labels used in the downstream tasks are not present in our pre-training data.

We follow the original data split: 18349 rooms for training, 1776 rooms for validation, and 1691 rooms for testing. We use {\SST} as the encoder and a simple decoder to output semantic labels of input points. The decoder is similar to the UNet decoder. We upsample the features from the coarsest level using interpolation, followed by a Linear Layer to match the dimensions. We then add fine-level features from the encoder using skip-connection. The purpose of the simple decoder is to make the backbone the main factor in feature learning. The network was trained with 100 epochs, a batch size of 12, and augmented input data through random cropping and rotation. We use the AdamW optimizer with a Cosine learning rate scheduler. \cref{tab:perf-nets} reports the network parameters, the amortized inference latency measured in the Structured3D validation set, and the average and peak memory footprint of network training in a subset (600 samples) of the training dataset. Training \SST-S and \SST-L took 488 and 703 GPU hours with NVidia V100 GPUs, respectively. We also pretrained {\SST}$_n$-S and {\SST}$_n$-L and use them only for the ScanNet segmentation task.
 During the network training phase, we employed the data cropping strategy of \cite{lai2022stratified} to randomly crop a portion from the input scene for training, with a maximum of \num{120000} sparse voxels. Additionally, we use the data augmentation technique of \cite{lai2022stratified} for network training.

 \begin{table}[t]
    \centering
    \resizebox{\columnwidth}{!}{
        \begin{tabular}{lccc}
            \toprule
            \mythead{Model} & \mythead{Params} & \mythead{Latency} & \mythead{Memory} (Avg./Peak) \\
            \midrule
            {\SST}-S      & \SI{23.57}{M}  & \SI{377.98}{ms} & 2.24/3.69 GB               \\
            {\SST}-L      & \SI{60.75}{M}  & \SI{554.58}{ms} & 4.11/6.73 GB               \\
            \bottomrule
        \end{tabular}
    }
    \vspace{2pt}
    \caption{Model parameters, inference latency, and memory footprint, evaluated on Structured3D segmentation.}
    \label{tab:perf-nets}  \vspace{-2mm}
\end{table}
\myparagraph{Fine-tuning for downstream tasks} We employ our pretrained backbone as a multi-resolution feature encoder and attach it to a task-specific decoder for downstream tasks. The pretrained network weights and look-up tables are loaded for initialization, while the decoder weight is randomly initialized. Our experiments on downstream tasks are presented in \cref{sec:results}.

\section{Experimental Analysis} \label{sec:results}

We conducted experiments to assess the scalability, generality, and effectiveness of our backbone design for typical indoor scene understanding tasks. This section is structured as follows: we first present the experimental setup of downstream tasks in \cref{subsec:setting}, then evaluate our model's capability through a series of experiments, comparisons, and ablation studies in \cref{subsec:eval_task,subsec:eval_scale,subsec:ablation}.

\subsection{Experimental settings for downstream tasks} \label{subsec:setting}

\subsubsection{Semantic segmentation}
\myparagraph{Datasets} ScanNet~\cite{dai2017scannet} contains 1613 indoor scans with 20 common semantic labels. We followed the original training/validation/test data split and reported the mean Intersection over Union (IoU) on the validation and test datasets. S3DIS contains 272 rooms from 6 large-scale areas. One area was selected as the validation set, and the other areas were used for training. We reported mean IoU on Area-5 and 6-fold cross-validation results.

\myparagraph{Networks}
To adapt our pretrained encoders to the segmentation task, we revised the decoder structure used for our pretraining by adding a {\SST} block at each level after the skip-connection to improve the decoder capability. The minimum voxel size for ScanNet and S3DIS is \SI{2}{cm} and \SI{5}{cm}, respectively. As ScanNet point clouds possess point normal information and many existing methods utilized it, we used \SST$_n$-S and \SST$_n$-L for this task.

\myparagraph{Training}
 During the training stage, we augmented the data through random cropping, scaling, and rotation. We fine-tuned our models (\SST-S and \SST-L) for 600 epochs for both ScanNet and S3DIS, with a batch size of 12.

\subsubsection{3D detection}

\myparagraph{Datasets}
The ScanNet~\cite{dai2017scannet} dataset contains labels for 18 objects, with 1201 scans used for training and 312 for validation. The S3DIS~\cite{S3DIS} dataset contains instance labels of 7 categories, including floor and ceiling. Following FCAF3D~\cite{rukhovich2021fcaf3d}, we excluded these two categories and trained and validated our models on the remaining five categories. AP@0.25 and AP@0.5 are the evaluation metrics employed in our experiment.

\myparagraph{Networks}
We replaced the encoders of two state-of-the-art 3D detection networks (FCAF3D~\cite{rukhovich2021fcaf3d} and CAGroup3D~\cite{wang2022cagroupd}) with our pretrained {\SST} to demonstrate the advantages of using our pretrained backbones for 3D detection. Specifically, For FCAF3D, we replaced the Sparse-Convolution-based ResNet in FCAF3D with our pretrained {\SST} and kept the other modules unchanged, which we named {\SST}-S+FCAF3D and {\SST}-L+FCAF3D; for CAGroup3D, we replaced its feature extractor (BiResNet) with our pretrained encoder with upsample layers, and kept the other modules for proposal generation and the detection head unchanged, which we named {\SST}-S+CAGroup3D and {\SST}-L+CAGroup3D.

 \myparagraph{Training}
For network training, we set the finest voxel size to \SI{2}{cm} for both ScanNet and S3DIS, and use the same data augmentation as CAGroup3D and FCAF3D. We adjusted our \SST-S+CAGroup3D and \SST-S+FCAF3D models for ScanNet 3D detection by training them for 200 epochs with a batch size of 12. Similarly, for S3DIS 3D detection, we fine-tuned our \SST-L+FCAF3D model for 200 epochs with a batch size of 8.

\subsection{Performance on downstream tasks} \label{subsec:eval_task}

\begin{table*}[t]
    \centering
    \resizebox{0.8\linewidth}{!}{
    \scriptsize
    \begin{tabular}{c|cc|cc}
        \toprule
        \multirow{2}{*}{\mythead{Method}}               & \multicolumn{2}{c|}{\mythead{ScanNet Segmentation}} & \multicolumn{2}{c}{\mythead{S3DIS Segmentation}}                                       \\
                                                        & \mythead{Val mIoU}  (\%)                                    & \mythead{Test mIoU}  (\%)                                  & \mythead{Area5 mIoU}  (\%) & \mythead{6-fold mIoU} (\%) \\
        \midrule
        FastPointTransformer~\cite{park2022fast}        & 72.4                                                & -                                                & 70.4             & -                \\
        PointMetaBase-XXL~\cite{lin2022meta}            & 72.8                                                & 71.4                                             & 72.0             & 77.0             \\
        LargeKernel3D~\cite{chen2022scaling}            & 73.5                                                & 73.9                                             & -                & -                \\
        Mix3D~\cite{nekrasov2021mix3d}                  & 73.6                                                & \textbf{78.1}                                    & -                & -                \\
        Stratified Transformer~\cite{lai2022stratified} & 74.3(73.1)                                          & 74.7                                             & 72.0             & -                \\

        PointConvFormer~\cite{wu2022pointconvformer}    & 74.5                                                & 74.9                                             & -                & -                \\
        O-CNN~\cite{Wang2017}                           & 74.5                                                & 76.2                                             & -                & -                \\
        PointTransformerV2~\cite{wu2022point}           & 75.4(74.4)                                          & 75.2                                             & 71.6             & -                \\
        OctFormer~\cite{octformer}                      & 75.7(74.5)                                          & 76.6                                             &                  & -
        \\
        PointNeXt-XL~\cite{qian2022pointnext}           & -                                                   & -                                                & 71.1             & 74.9             \\
        PointMixer~\cite{choe2021pointmixer}            & -                                                   & -                                                & 71.4             & -                \\
        WindowNorm~\cite{wang2022window}                & -                                                   & -                                                & 72.2             & \textbf{77.6}    \\
        \rowcolor{gray!20} {\SST}-S$^\star$             & 75.5(74.5)                                          & -                                                & \textbf{72.5}    & 76.9             \\
        \rowcolor{gray!20} {\SST}$_n$-S$^\star$         & \textbf{76.4}(75.2)                                          & -                                                & -                & -                \\



        \midrule
            DepthContrast~\cite{zhang2021self}              & 71.2                                                & -                                                & 70.6             & -                \\
        SceneContext~\cite{hou2020efficient}            & 73.8                                                & -                                                & 72.2             & -                \\
       PointContrast~\cite{Xie2020}                    & 74.1                                                & -                                                & 70.9             & -                \\
        MaskContrast~\cite{wu2023masked}                & 75.5(74.4)                                          & -                                                & -                & - \\
        \rowcolor{gray!20} {\SST}-S                     & 76.1(75.0)                                          & -                                                & 73.0             & 78.2             \\
        \rowcolor{gray!20} {\SST}$_n$-S                 & 76.8(75.7)                                          & -                                                & -                & -                \\

        \rowcolor{gray!20} {\SST}-L                     & 76.7(75.7)                                          & -                                                & \textbf{74.5}    & \textbf{79.8}    \\

        \rowcolor{gray!20} {\SST}$_n$-L                 & \textbf{77.5}(76.4)                                 & 77.9                                             & -                & -                \\
        \bottomrule
    \end{tabular}
    }
    \vspace{2pt}
    \caption{Quantitative evaluation on semantic segmentation. The methods in the upper part of the table are supervised methods, while those in the lower part are based on pre-training. We present the highest scores achieved by the compared approaches. 
        On the ScanNet benchmark (test dataset), we ensembled the results of three trained models by voting the prediction on over-segmented meshes, similar to Mix3D.  The paper of WindowNorm is a concurrent and unpublished work.
    }  \label{tab:seg-all-in-one} \vspace{-2mm}
\end{table*}

\subsubsection{Semantic segmentation}\label{subsubsec:exp_segmentation}
We quantitatively evaluated our pretrained models with fine-tuning on the semantic segmentation task and compared them to state-of-the-art methods, as shown in \cref{tab:seg-all-in-one}. Following existing approaches such as StratifiedTransformer~\cite{lai2022stratified}, PointransformerV2~\cite{wu2022point}, O-CNN~\cite{Wang2017}, OctFormer~\cite{octformer} that perform test-time augmentation in evaluating segmentation performance, we voted our segmentation results via 12 rotation augmentation. For reference, we also report the unvoted results of several methods, including ours. The unvoted results are reported in parentheses.
In the following, we analyze the results in detail.

\paragraph{Comparison with supervised methods}
Among the existing supervised methods, PointTransformerV2~\cite{wu2022point}, O-CNN~\cite{Wang2017}, OctFormer~\cite{octformer} utilize point normals on ScanNet segmentation.
Our \SST-S outperforms the best supervised method by 0.4 mIoU on ScanNet val, +0.8 mIoU on S3DIS Area5, and +0.6 mIoU on S3DIS 6-fold. With greater capacity, our \SST-L surpasses all the compared supervised methods with 1.0 mIoU on ScanNet val, +2.3 mIoU on S3DIS Area5, and +2.2 mIoU on S3DIS 6-fold. With point normal inputs for pre-training and fine-tuning, our \SST$_n$-L further enhances performance on ScanNet val by +0.8 mIoU. We also present the results of \SST$_n$-S on ScanNet, which is +0.7 mIoU better than \SST-S.

\paragraph{Comparison with pretraining methods}
We compared our approach to existing unsupervised pretraining methods, such as PointContrast~\cite{Xie2020}, SceneContext~\cite{hou2020efficient},  DepthContrast~\cite{zhang2021self} and MaskContrast~\cite{wu2023masked}, which use multi-view data from ScanNet for pretraining and Minkowski U-Net~\cite{choy20194d} as the encoder architecture.  MaskConstrast also combines the ArkitScenes dataset~\cite{ARKitScenes} with ScanNet for pretraining. As can be seen in Table~\ref{tab:seg-all-in-one} (lower section), these unsupervised pretrained methods have lower performance than state-of-the-art supervised methods, with large gaps to our approach.  One may wonder if unsupervised pretrained methods can take advantage of large datasets. Our initial tests (see \cref{subsec:eval_scale}) indicate that the advantages are restricted, probably because the convolutional encoder structure is not able to effectively extract data priors in comparison to the transformer structure.

\begin{figure}[t]
    \centering
    \includegraphics[width=0.95\columnwidth]{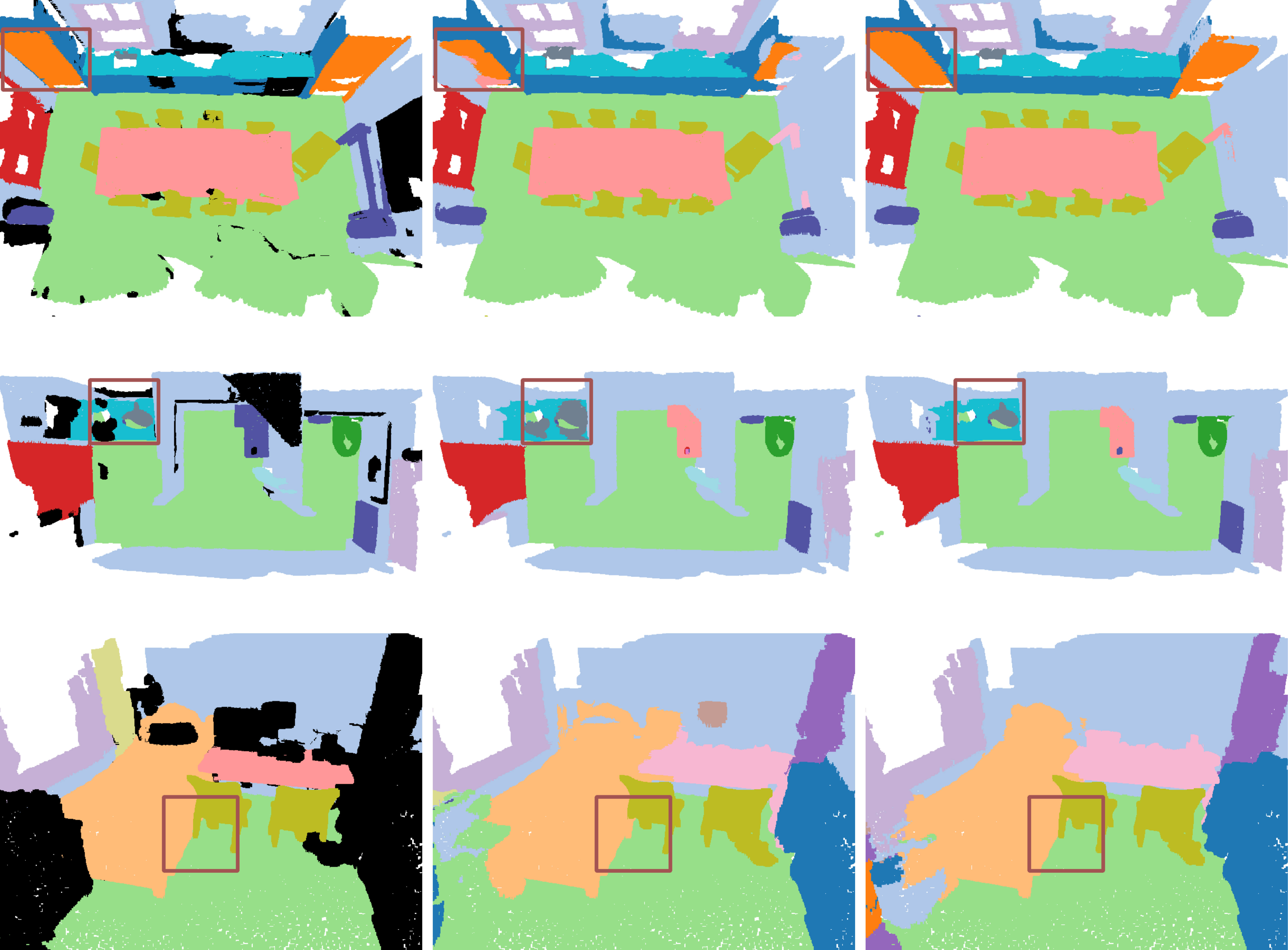}
    \caption{Visual comparison of ScanNet segmentation. \textbf{Left}: Ground truth segmentation labels (points color in black are not labeled in the original dataset). \textbf{Middle}: Stratified Transformer's results. \textbf{Right}: {\SST}$_n$-L's results.  
    }
    \label{fig:scannet_seg_vis}
\end{figure}
\begin{table*}[t]
    \centering
    \resizebox{1\linewidth}{!}{
        \scriptsize
        \setlength\tabcolsep{2.5pt}
        \begin{tabular}{c|c|cccccccccccccccccccc}
            \toprule
            \mythead{Method}                                & \mythead{mIoU}        &
            \rotatebox{90}{wall}                            & \rotatebox{90}{floor} & \rotatebox{90}{cabinet} & \rotatebox{90}{bed} & \rotatebox{90}{chair} & \rotatebox{90}{sofa} & \rotatebox{90}{table} & \rotatebox{90}{door} & \rotatebox{90}{window} & \rotatebox{90}{bookshelf} & \rotatebox{90}{picture} & \rotatebox{90}{counter} & \rotatebox{90}{desk} & \rotatebox{90}{curtain} & \rotatebox{90}{refri.} & \rotatebox{90}{shower cur.} & \rotatebox{90}{toilet} & \rotatebox{90}{sink} & \rotatebox{90}{bathtub} & \rotatebox{90}{other.}                                                 \\
            \midrule
            Stratified Transformer~\cite{lai2022stratified}            & 74.3                    & 86.2                & 95.1                  & 66.4                 & 80.9                  & 90.0                 & 82.9                   & 76.6                      & 69.0                    & 71.7                    & 82.2                 & 30.1                    & 66.0                   & 71.1               & 75.8                   & 66.5                 & 68.6                    & 94.6                   & 67.6          & 88.3          & 56.3          \\
            PointTransformerV2~\cite{wu2022point}                      & 75.4                    & 86.1                & \textbf{95.4}         & 67.4                 & \textbf{81.9}         & 91.9        & \textbf{86.5}          & 77.5            & 68.8                    & 68.7                    & 84.5                 & 34.1                    & 66.7          & 71.1               & 78.7                   & 69.2                 & 71.2                    & 94.5                   & 65.6          & \textbf{89.1} & 60.5          \\
            \midrule
            \rowcolor{gray!20} {\SST}$_n$-S & 76.8 & \textbf{87.6} & 95.1 & 67.0 & 79.3 & 91.6 & 83.7 & 78.2 & 69.6 & 71.0 & 86.8 & \textbf{39.9} & 68.1 & \textbf{73.6} & 80.2 & 72.6 & 75.8 & \textbf{95.6} & \textbf{71.2} & \textbf{89.7} & 60.2 \\

            \rowcolor{gray!20} {\SST}$_n$-L & \textbf{77.5} & 87.0 & 95.0 & \textbf{73.2} & 81.5 & \textbf{92.1} & 84.2 & \textbf{79.1} & \textbf{70.6} & \textbf{72.8} & \textbf{87.3} & 36.9 & \textbf{68.3} & 73.3 & \textbf{80.9} & \textbf{73.4} & \textbf{77.8} & 95.4 & 70.9 & 88.6 & \textbf{60.8} \\

            \bottomrule
        \end{tabular}
    }
    \vspace{2pt}
    \caption{Category-wise segmentation results evaluated on ScanNet validation set.}  \label{tab:sup_seg_scannet} 
\end{table*}

\begin{table}[t]
    \centering
    \resizebox{0.90\linewidth}{!}{
        \scriptsize
        \setlength\tabcolsep{2.5pt}
        \begin{tabular}{c|c|*{6}{c}}
            \toprule
            \mythead{Method}                      & \mythead{6-fold mIoU} &
            A1                                  & A2           & A3                  & A4            & A5   & A6                                                            \\
            \midrule
            \rowcolor{gray!20} {\SST}-S           & 78.2                & \textbf{82.2} & 64.2 & 83.7          & 75.9          & 73.0          & 86.3          \\
            \rowcolor{gray!20} {\SST}-L            & \textbf{79.8}       & 82.1          & 66.3 & \textbf{85.9} & \textbf{76.7} & \textbf{73.4} & \textbf{87.2} \\
            \bottomrule
        \end{tabular}
    }
    \vspace{2pt}
    \caption{6-fold S3DIS segmentation results. ``A$n$'' means that the $n$-th Area is the test data and other 5 Areas are used for training. The reported number is mIoU.  
    }  \label{tab:sup_seg_s3dis} 
\end{table}

\paragraph{Non-pretrained \SST}
We also evaluated our backbone structure without pretraining, \ie training {\SST} from scratch for downstream tasks.
e also trained \SST-S from scratch for comparison (600 epochs for ScanNet and 3000 epochs for S3DIS), denoted by \SST-S$^\star$ and \SST$_n$-S$^\star$. \SST-S$^\star$ has a comparative performance to existing works, only inferior to WindowNorm~\cite{wang2022window} on S3DIS 6-fold. \SST$_n$-S$^\star$ can further improve \SST-S$^\star$ due to the use of point normals. Despite their good performance, they are consistently inferior to those of their pre-trained versions, demonstrating the effectiveness of pretraining.

We noticed that \SST-L trained from scratch does not produce excellent results. \SST-L$^\star$ and \SST$_n$-L$^\star$ only achieved $73.9$ mIoU and $74.2$ mIoU on ScanNet segmentation (val), respectively. This is due to the overfitting caused by the large model size in comparison to the size of the training data (ScanNet or S3DIS). This is in agreement with previous findings in the NLP and vision fields (\eg Bert, SwinTransformerV2) that large transformer models need a considerable amount of data for pretraining to show their advantages.

\paragraph{Additional results}
In \cref{tab:sup_seg_scannet} and \cref{tab:sup_seg_s3dis}, we further provide semantic segmentation performance reports on all categories of ScanNet and S3DIS. For 6-fold cross validation of S3DIS, we fine-tuned our models for 600 epochs only. And in the test phase, we voted our predictions across 12 rotation augmentations. As for Area 5, we can achieve 74.5 mIoU (voted) by fine-tuning our model for 3000 epochs.  For visualization, we illustrate three segmentation results in \cref{fig:scannet_seg_vis}.

\subsubsection{3D detection} \label{subsubsec:exp_det}
We quantitatively evaluated our pretrained models with fine-tuning on the 3D detection task and compared them to state-of-the-art methods, as shown in \cref{tab:detect-scannet,tab:detect-s3dis}. In the following, we analyze the results in detail.

\begin{table}[t]
    \centering
    \resizebox{0.8\linewidth}{!}{
    \scriptsize
    \begin{tabular}{l|cc}
        \toprule
        \mythead{Method}                      & \mythead{mAP@0.25} & \mythead{mAP@0.5} \\
        \midrule
        RepSurf~\cite{ran2022surface}         & 71.2               & 54.8              \\
        FCAF3D~\cite{rukhovich2021fcaf3d}     & 71.5               & 57.3              \\
        SoftGroup~\cite{vu2022softgroup}      & 71.6               & 59.4              \\
        CAGroup3D~\cite{wang2022cagroupd}     & \textbf{75.1}      & \textbf{61.3}     \\
        \rowcolor{gray!20} {\SST}-S$^\star$+FCAF3D    & 72.1               & 56.8              \\
        \rowcolor{gray!20}{\SST}-S$^\star$+CAGoup3D   & 73.3               & 58.6              \\
        \midrule
        Point-M2AE~\cite{zhang2022point}      & 50.1               & 33.2              \\
        PointContrast~\cite{Xie2020}          & 59.2               & 37.3              \\
        RandomRooms~\cite{rao2021randomrooms} & 68.6               & 51.5              \\
        \rowcolor{gray!20}{\SST}-S+FCAF3D     & 74.2               & 59.5              \\
        \rowcolor{gray!20}{\SST}-L+FCAF3D     & 74.2               & 58.6              \\
        \rowcolor{gray!20}{\SST}-S+CAGoup3D   & 76.4               & 62.7              \\
        \rowcolor{gray!20}{\SST}-L+CAGoup3D   & \textbf{76.4}      & \textbf{63.2}     \\
        \bottomrule
    \end{tabular}
    }
    \vspace{2pt}
    \caption{Quantitative evaluation on 3D detection (ScanNet). The methods in the upper part of the table are supervised methods, while those
in the lower part are based on pretraining.}  \label{tab:detect-scannet} 
\end{table}

\begin{table}[t]
    \centering
    \resizebox{0.9\linewidth}{!}{
        \scriptsize
        \begin{tabular}{l|cc}
            \toprule
            \mythead{Method}                           & \mythead{mAP@0.25} & \mythead{mAP@0.5} \\
            \midrule
            GSDN~\cite{gwak2020generative}             & 47.8               & 25.1              \\
            FCAF3D~\cite{rukhovich2021fcaf3d}          & \textbf{66.7}      & \textbf{45.9}     \\
            \rowcolor{gray!20} {\SST}-S$^\star$+FCAF3D         & 64.6               & 40.7              \\
            \midrule
            \rowcolor{gray!20}{\SST}-S+FCAF3D          & 69.9               & 50.2              \\
            \rowcolor{gray!20}{\SST}-L+FCAF3D          & 72.1               & 54.0              \\
            \rowcolor{gray!20}{\SST}-S+FCAF3D(2-stage) & \textbf{75.4}      & \textbf{58.6}     \\
            \bottomrule
        \end{tabular}
    }
    \vspace{2pt}
    \caption{Quantitative evaluation on 3D detection (S3DIS). 
        }  \label{tab:detect-s3dis}
\end{table}

\begin{table*}[t]
    \centering
    \resizebox{\linewidth}{!}{
    \scriptsize
    \setlength\tabcolsep{2.5pt}
    \begin{tabular}{c|c|c|cccccccccccccccccc}
        \toprule
        \mythead{Method}                        & \mythead{Pre.}        & \mythead{mAP@0.5}       &
        \rotatebox{90}{cabinet}               & \rotatebox{90}{bed} & \rotatebox{90}{chair} & \rotatebox{90}{sofa} & \rotatebox{90}{table} & \rotatebox{90}{door} & \rotatebox{90}{window} & \rotatebox{90}{bookshelf} & \rotatebox{90}{picture} & \rotatebox{90}{counter} & \rotatebox{90}{desk} & \rotatebox{90}{curtain} & \rotatebox{90}{refri.} & \rotatebox{90}{shower cur.} & \rotatebox{90}{toilet} & \rotatebox{90}{sink} & \rotatebox{90}{bathtub} & \rotatebox{90}{other.}                                                 \\
        \midrule
        FCAF3D\cite{rukhovich2021fcaf3d}      & \xmark              & 57.3                  & 35.8                 & 81.5                  & 89.8                 & 85.0                   & 62.0                      & 44.1                    & 30.7                    & 58.4                 & 17.9                    & 31.3                         & 53.4                           & 44.2                   & 46.8                 & \textbf{64.2}           & 91.6                           & 52.6          & 84.5          & 57.1          \\
        CAGroup3D\cite{wang2022cagroupd}      & \xmark              & 61.3                  & 41.4                 & 82.8                  & 90.8                 & \textbf{85.6}          & 64.9                      & 54.3                    & 37.3                    & 64.1                 & 31.4                    & 41.1                         & \textbf{63.6}                  & 44.4                   & 57.0                 & 49.3                    & 98.2                           & \textbf{55.4} & 82.4          & 58.8          \\
        \midrule
        \rowcolor{gray!20} {\SST}-S+CAGroup3D & \checkmark          & 62.7                  & 45.7                 & 82.7                  & \textbf{91.0}        & 79.5                   & \textbf{67.4}             & \textbf{57.5}           & \textbf{42.7}           & \textbf{59.8}        & 36.8                    & 40.4                         & 62.6                           & \textbf{48.2}          & \textbf{60.7}        & 59.8                    & \textbf{99.7}                  & 55.1          & 77.6          & 60.7          \\
        \rowcolor{gray!20} {\SST}-L+CAGroup3D & \checkmark          & \textbf{63.2}         & \textbf{46.1}        & \textbf{85.5}         & \textbf{91.0}        & 81.1                   & 64.5                      & 52.9                    & 42.4                    & 57.8                 & \textbf{38.2}           & \textbf{47.2}                & 63.4                           & 46.0                   & 59.3                 & 61.7                    & 98.3                           & 54.2          & \textbf{85.4} & \textbf{63.6} \\
        \bottomrule
    \end{tabular}
    }
    \vspace{2pt}
    \caption{ Quantitative comparison on 3D detection (ScanNet).}  \label{tab:sup_det_scannet} 
\end{table*}

\begin{table}[t]
    \centering
    \resizebox{0.95\linewidth}{!}{
    \scriptsize
    \setlength\tabcolsep{2.5pt}
    \begin{tabular}{c|c|c|ccccc}
        \toprule
        \mythead{Method}                   & \mythead{Pre.} & \mythead{mAP@0.5} &
        table                            & chair        & sofa            & bookcase      & board                                                        \\
        \midrule
        GSDN \cite{gwak2020gsdn}         & \xmark       & 25.1            & 36.6          & 75.3          & \;\;6.1           & \;\;6.5           & 1.2          \\
        FCAF3D\cite{rukhovich2021fcaf3d} & \xmark       & 45.9            & 45.4          & 88.3          & 70.1          & 19.5          & 5.6          \\
        \midrule
        \rowcolor{gray!20} {\SST}-S      & \checkmark   & 50.2            & 52.8          & \textbf{90.4} & 78.8          & \textbf{20.9} & 7.9          \\
        \rowcolor{gray!20} {\SST}-L      & \checkmark   & \textbf{54.0}   & \textbf{56.2} & 90.3          & \textbf{95.1} & 19.6          & \textbf{8.9} \\
        \bottomrule
    \end{tabular}
    }
    \vspace{2pt}
    \caption{Quantitative comparison on 3D detection (S3DIS).}  \label{tab:sup_det_s3dis} 
\end{table}

\begin{figure}[t]
    \centering
    \includegraphics[width=0.95\columnwidth]{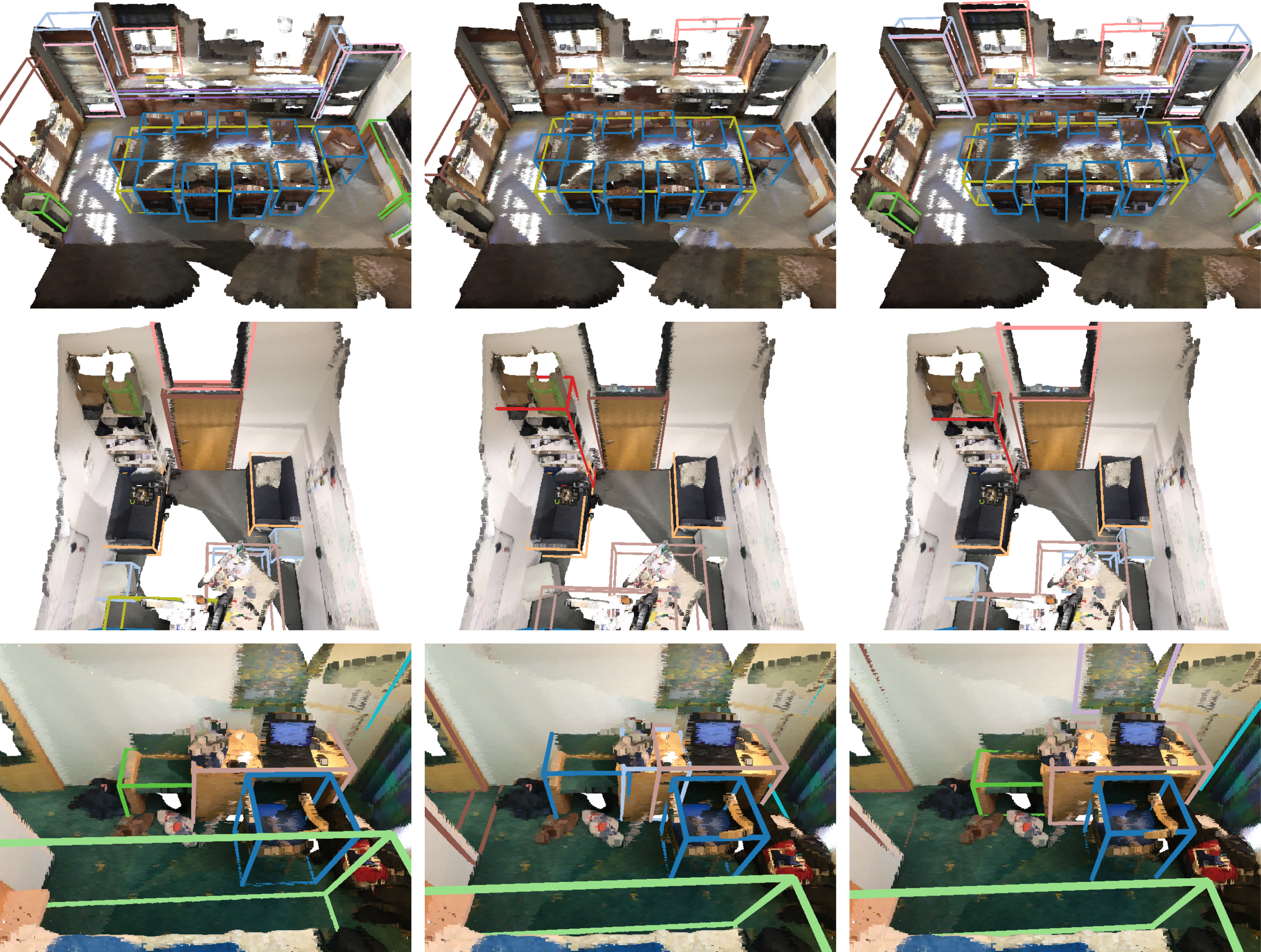}
    \caption{Visual comparison of ScanNet 3D detection. \textbf{Left}: Ground truth. \textbf{Middle}: FCAF3D's result. \textbf{Right}: {\SST}-L+CAGroup3D's results. Note that the original CAGroup3D paper does not release its checkpoint, so no visual results provided in this figure.}
    \label{fig:scannet_det_vis}
\end{figure}

\begin{figure}[t]
    \centering
    \includegraphics[width=0.95\columnwidth]{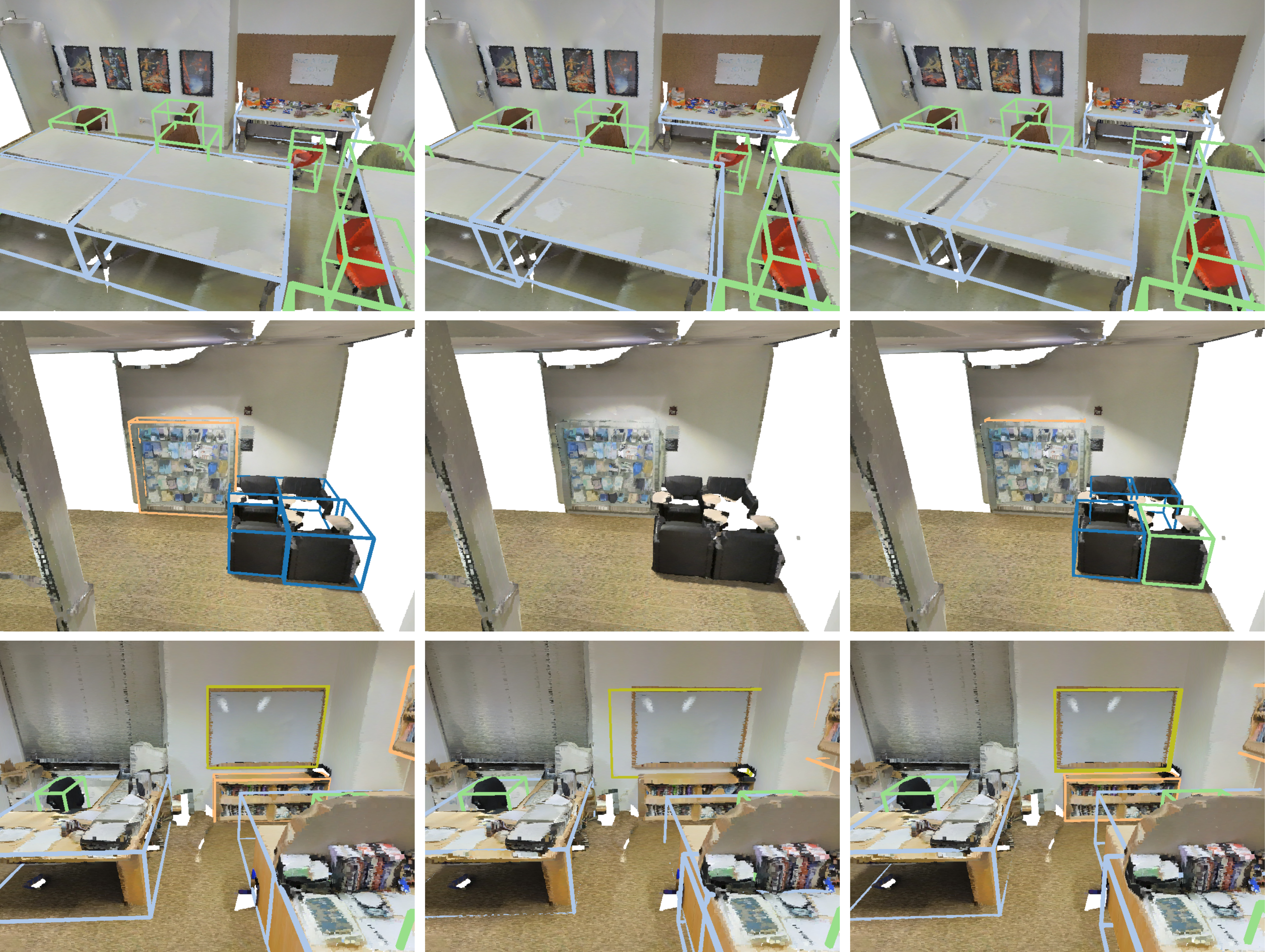}
    \caption{Visual comparison on S3DIS 3D detection. \textbf{Left}: Ground-truth 3D bounding boxes. \textbf{Middle}: FCAF3D's detection result. \textbf{Right}: {\SST}-L+FCAF3D's detection results. Our {\SST-L}+FCAF3D generates more compact and accurate proposals of table, sofa, and board, which are consistent with the result shown in \cref{tab:sup_det_s3dis}.}
    \label{fig:s3dis_det_vis}
\end{figure}

\paragraph{Comparison with supervised methods}
On ScanNet 3D detection (as shown in Table \ref{tab:detect-scannet}), {\SST}-S+FCAF3D and {\SST}-L+FCAF3D both outperform FCAF3D, with an increase of 2.7 points in mAP@0.25. {\SST}-S+CAGroup3D and {\SST}-L+CAGroup3D both surpass CAGroup3D, with {\SST}-L+CAGroup3D achieving a new record on mAP@0.50. Table \ref{tab:detect-s3dis} shows the performance of our model on S3DIS. Compared to FCAF3D, our pretrained backbones significantly improve the performance, with {\SST}-S+FCAF3D increasing by 4.3 points and {\SST}-L+FCAF3D increasing by 8.1 points in mAP@0.5.
The experiments of detection lead us to the conclusion that semantic segmentation is an effective pretext task for 3D pretraining, as our pretrained backbone demonstrates remarkable generality to the 3D detection task.

\myparagraph{Two-stage finetuning} We discovered through experimentation that S3DIS detection can be improved by a two-step finetuning process. Initially, we fine-tuned {\SST}-S on ScanNet detection. Then, we loaded the fine-tuned {\SST} and continued the fine-tuning on S3DIS training data, taking advantage of the prior knowledge gained from real data in the first step. The improved performance is reported in \cref{tab:detect-s3dis}.

\paragraph{Comparison with pretrained methods}
We compared our approach to three existing unsupervised pretraining methods: PointContrast~\cite{Xie2020}, Point-M2AE~\cite{zhang2022point} and RandomRooms~\cite{rao2021randomrooms}. PointContrast employed ScanNet as its pretraining dataset and Minkowski U-Net as its encoder, Point-M2AE utilized both ShapeNet~\cite{shapenet2015} and ScanNet as its pretraining dataset and a transformer encoder, and RandomRooms adopted random samples of multiple objects from ShapeNet as its pretraining dataset and PointNet++ as its backbone. In line with the observations in semantic segmentation, these unsupervised pretraining methods were unable to compete with supervised methods and our approaches.

\paragraph{Non-pretrained \SST}
Without pretraining, our model \SST-S$^\star$ that were trained from scratch show average results, however, they are not as good as the state-of-the-art methods. We speculate that more training data for these tasks is required for training our transformer from scratch. The higher performance of our pretrained \SST ~also reflect this fact.

\myparagraph{Additional results}
In \cref{tab:sup_det_scannet,tab:sup_det_s3dis}, we present the category-wise performance report on ScanNet and S3DIS detection. Visualizations of some 3D detection results can be seen in \cref{fig:scannet_det_vis,fig:s3dis_det_vis}.

\subsection{Model scalability} \label{subsec:eval_scale}
Our backbone design takes advantage of large pretrained datasets, particularly for our large model {\SST}-L. To evaluate the impact of the amount of pretrained data on the performance of downstream tasks with respect to backbone architectures, we conducted the following experiments.

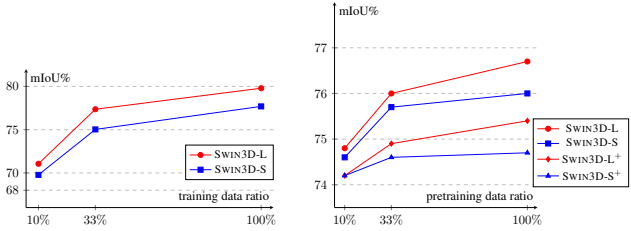
\begin{figure}[t]
    \centering
    \begin{tikzpicture}[scale=0.48]
        \begin{axis}[
                unit vector ratio=2 7 1,
                xlabel={training data ratio},
                ylabel={mIoU\%},
                axis line style={->},
                axis lines=middle,
                xmin=5, xmax=105,
                ymin=66, ymax=82,
                xticklabel={$\pgfmathprintnumber{\tick}\%$},
                xtick={10,33,100},
                ytick={68,70,75,80},
                legend style={at={(1,0.2)},anchor=south east},
                ymajorgrids=true,
                grid style=dashed,
            ]
            \addplot[
                color=red,
                mark=*,
            ]
            coordinates {
                    (10,71.06)(33,77.36)(100,79.79)
                };

            \addplot[
                color=blue,
                mark=square*,
            ]
            coordinates {
                    (10, 69.78)(33,75.03)(100,77.69)
                };

            \legend{\small{\SST-L},\small{\SST-S}}
        \end{axis}
    \end{tikzpicture}
    \hfill
    \begin{tikzpicture}[scale=0.48]
        \begin{axis}[
                unit vector ratio=2 45 1,
                xlabel={pretraining data ratio},
                ylabel={mIoU\%},
                axis line style={->},
                axis lines=middle,
                xticklabel={$\pgfmathprintnumber{\tick}\%$},
                xmin=5, xmax=105,
                ymin=73.5, ymax=78,
                xtick={10,33,100},
                ytick={74,75,76,77},
                legend style={at={(1.45,0.1)},anchor=south east},
                ymajorgrids=true,
                grid style=dashed,
            ]

            \addplot[
                color=red,
                mark=*,
            ]
            coordinates {
                    (10, 74.8)(33,76.0)(100,76.7)
                };

            \addplot[
                color=blue,
                mark=square*,
            ]
            coordinates {
                    (10, 74.6)(33,75.7)(100,76.0)
                };

            \addplot[
                color=red,
                mark=diamond*,
            ]
            coordinates {
                    (10, 74.2)(33,74.9)(100,75.4)
                };

            \addplot[
                color=blue,
                mark=triangle*,
            ]
            coordinates {
                    (10, 74.2)(33,74.6)(100,74.7)
                };

            \legend{\small{\SST-L},\small{\SST-S},\small{\SST-L$^+$},\small{\SST-S$^+$}}
        \end{axis}
    \end{tikzpicture}

    \caption{Model scalability with respect to different ratios of data for pretraining. \textbf{Left}: Test on Structured3D segmentation. \textbf{Right}: \SST-L on downstream ScanNet segmentation. \SST-L$^+$ and \SST-S$^+$ stand for the models pretrained and fine-tuned with cRPE instead of cRSE.}
    \label{fig:modelscale} 
\end{figure}

\paragraph{Scalability of {\SST}} We pretrained our backbones with different amounts of training data: 10\%, 33\%, and 100\%. We then fine-tuned them for downstream tasks. \cref{fig:modelscale} shows the segmentation performance of our {\SST}-S and {\SST}-L models on the test dataset of Structured3D (left) and the validation set of ScanNet segmentation (right). The plots demonstrate that (1) with more training data, the performance of both models increases significantly; (2) {\SST}-L has more capacity to benefit from large data and performs better than {\SST}-S.

\paragraph{Scalability on the use of cRSE}
We pretrained our backbones with cRPE instead of our cRSE, referred to as {\SST}-S$^+$ and {\SST}-L$^+$. We then fine-tuned them on the ScanNet segmentation task, and the results are shown in the right side of \cref{fig:modelscale}. The plots show clearly that pretraining with our cRSE scheme exhibits better scalability than using cRPE. We also found that \SST-S
outperforms {\SST}-L$^+$, which indicates that the capture of irregular signals is essential for backbone learning.

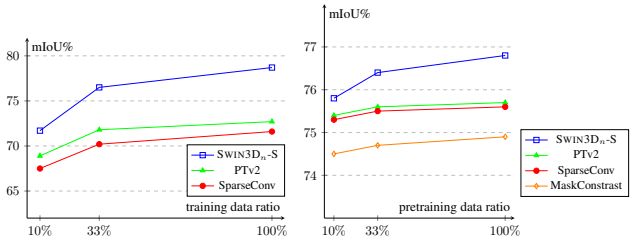
\begin{figure}[t]
    \centering

    \begin{tikzpicture}[scale=0.5]
        \begin{axis}[
                unit vector ratio=2 7 45,
                xlabel={training data ratio},
                ylabel={mIoU\%},
                axis line style={->},
                axis lines=middle,
                xmin=5, xmax=105,
                ymin=62, ymax=82,
                xticklabel={$\pgfmathprintnumber{\tick}\%$},
                xtick={10,33,100},
                ytick={65,70,75,80},
                legend style={at={(1,0.12)},anchor=south east},
                ymajorgrids=true,
                grid style=dashed,
            ]
            \addplot[
                color=blue,
                mark=square,
            ]
            coordinates {
                    (10, 71.70)(33,76.50)(100,78.70)
                };
            \addplot[
                color=green,
                mark=triangle*,
            ]
            coordinates {
                    (10, 68.90)(33,71.80)(100,72.70)
                };
            \addplot[
                color=red,
                mark=*,
            ]
            coordinates {
                    (10,67.50)(33,70.20)(100,71.60)
                };
            \legend{\small{\SST$_n$-S},\small{PTv2},\small{SparseConv}}
        \end{axis}
    \end{tikzpicture}
    \hfill
    \begin{tikzpicture}[scale=0.5]
        \begin{axis}[
                unit vector ratio=2 45 45,
                xlabel={pretraining data ratio},
                ylabel={mIoU\%},
                axis line style={->},
                axis lines=middle,
                xticklabel={$\pgfmathprintnumber{\tick}\%$},
                xmin=5, xmax=105,
                ymin=73, ymax=78,
                xtick={10,33,100},
                ytick={74,75,76,77},
                legend style={at={(1.6,0.1)},anchor=south east},
                ymajorgrids=true,
                grid style=dashed,
            ]
            \addplot[
                color=blue,
                mark=square,
            ]
            coordinates {
                    (10, 75.8)(33,76.4)(100,76.8)
                };
            \addplot[
                color=green,
                mark=triangle*,
            ]
            coordinates {
                    (10,75.4)(33,75.6)(100,75.7)
                };
            \addplot[
                color=red,
                mark=*,
            ]
            coordinates {
                    (10,75.3)(33,75.5)(100,75.6)
                };
            \addplot[
                color=orange,
                mark=diamond,
            ]
            coordinates {
                    (10,74.5)(33,74.7)(100,74.9)
                };
            \legend{\small{\SST$_n$-S},\small{PTv2},\small{SparseConv},\small{MaskConstrast}}
        \end{axis}
    \end{tikzpicture}
    \caption{We tested the scalability of different 3D backbones with respect to different ratios of data for pretraining. ``PTv2'' refers to PointTranformerV2. \textbf{Left}: The results for Structured3D segmentation. \textbf{Right}: The results for ScanNet segmentation (val). 
    }
    \label{fig:modelscale_backbone} 
\end{figure}

\paragraph{Scalability of other backbone architectures}
We selected a few representative backbone architectures, such as SparseConv Net used by MaskContrast~\cite{wu2023masked} and PointTransformerV2, and pre-trained them on the Structured3D dataset with a segmentation pretext task similar to ours. Additionally, we retrained MaskContrast in its unsupervised manner with the Structured3D dataset.  All these pretrained backbones were finetuned on the ScanNet segmentation task. Noted that for fair comparison, all these backbones used both color and normal signals for input in pre-training and fine-tuning stages. Their performance with respect to different amounts of pre-training data: 10\%, 33\%, and 100\% are plotted in \cref{fig:modelscale_backbone}. We discovered that these backbones had limited advantages from large datasets and had a large performance gap compared to our \SST-S, which has similar or fewer network parameters than theirs.
The results of this experiment demonstrate that simply increasing the amount of data used for pre-training does not necessarily lead to significant performance improvement for some existing 3D backbones.


\subsection{Ablation study} \label{subsec:ablation}

\myparagraph{Dataset for pretraining} We contrasted our backbones that had been pre-trained on either ScanNet or Structured3D for the downstream S3DIS segmentation task. \cref{tab:ablation-pretrain-dataset} clearly demonstrates that our backbones gain more from the extensive synthetic Structured3D data than from the real but limited ScanNet data. This implies that the size of the training data is the most critical factor in 3D backbone training, despite the disparity between real and synthetic data.

\myparagraph{Efficacy of cRSE} We conducted two experiments to assess the effectiveness of cRSE. In the first, we trained {\SST}-S from scratch in three different configurations: (1) using cRSE on point position only, similar to \cite{lai2022stratified}; (2) using cRSE on point position and color; (3) using cRSE on point position, color, and point normal. We also either excluded or included the normal information from the input point cloud for training and testing. The results in \cref{tab:ablation-crpe} demonstrate that taking into account color and normal variation via cRSE can significantly enhance the performance. In the second experiment, we used the pretrained \SST$_n$-L for ScanNet segmentation but did not load the pretrained look-up tables for color and/or normal components and trained these unloaded look-up tables from scratch during the network fine-tuning stage. The results in \cref{tab:ablation-pretrain-cpre} show that the use of pretrained tables is essential for improved performance.

\begin{table}[t]
    \centering
    \resizebox{\columnwidth}{!}{
        \begin{tabular}{lccc}
            \toprule
            \mythead{Backbone} & \mythead{Pre. dataset} & \mythead{Area5 mIoU} (\%) & \mythead{6-fold  mIoU} (\%) \\
            \midrule
            {\SST}-S         & ScanNet              & 71.8                    & 76.7                      \\
            {\SST}-S         & Structured3D         & \textbf{73.0}           & \textbf{78.2}             \\
            \midrule
            {\SST}-L         & ScanNet              & 68.9                    & 73.6                      \\
            {\SST}-L         & Structured3D         & \textbf{74.5}           & \textbf{79.8}             \\
            \bottomrule
        \end{tabular}
    }
    \vspace{2pt}
    \caption{S3DIS segmentation results by using the backbones pretrained on different datasets.}
    \label{tab:ablation-pretrain-dataset} \vspace{-2mm}
\end{table}

\begin{table}[t]
    \centering
    \scriptsize
    \begin{tabular}{llc}
        \toprule
        \mythead{Input Point Signal} & \mythead{cRSE}     & \mythead{Val mIoU} (\%) \\
        \midrule
        pos+color                  & pos              & 73.1                  \\
        pos+color                  & pos+color        & \textbf{74.5}         \\
        \midrule
        pos+color+normal           & pos              & 73.1                  \\
        pos+color+normal           & pos+color        & 74.4                  \\
        pos+color+normal           & pos+color+normal & \textbf{75.2}         \\
        \bottomrule
    \end{tabular}
    \vspace{2pt}
    \caption{Efficacy evaluation of cRSE on ScanNet segmentation. }
    \label{tab:ablation-crpe} 
\end{table}

\begin{table}[t]
    \centering
    \scriptsize
    \begin{tabular}{clc}
        \toprule
        \mythead{Backbone} & \mythead{loaded look-up tables} & \mythead{Val mIoU (\%)} \\
        \midrule
        {\SST}$_n$-L     & pos                           & 75.4                  \\
        {\SST}$_n$-L     & pos+color                     & 75.9                  \\
        {\SST}$_n$-L     & pos+color+normal              & \textbf{76.4}         \\
        \bottomrule
    \end{tabular}
    \vspace{2pt}
    \caption{Ablation study of using the pretrained look-up tables on ScanNet segmentation.  }
    \label{tab:ablation-pretrain-cpre} 
\end{table}

\section{Conclusion} \label{sec:conclusion}
In this paper, we introduce a novel 3D backbone --- {\SST} for indoor scene understanding, which demonstrates its scalability, generality, and superior performance through extensive experiments. We identify several possible directions for future research that can further exploit the potential of {\SST}. First, we would like to explore self-supervised pretraining methods using our backbone and enrich its representation with more real and synthetic data, including outdoor 3D data. Second, we propose to leverage image data and to integrate both pretrained image backbones and 3D backbones to enhance the effectiveness of 3D learning, as point clouds are often accompanied by high-resolution multiview images provided by 3D capture devices.

{\small
  \bibliographystyle{ieee_fullname}
  \bibliography{src/reference}

\begin{thebibliography}{10}\itemsep=-1pt

\bibitem{S3DIS}
Iro Armeni, Ozan Sener, Amir~Roshan Zamir, Helen Jiang, Ioannis~K. Brilakis,
  Martin Fischer, and Silvio Savarese.
\newblock
  \href{https://openaccess.thecvf.com/content_cvpr_2016/papers/Armeni_3D_Semantic_Parsing_CVPR_2016_paper.pdf}{3D
  semantic parsing of large-scale indoor spaces}.
\newblock In {\em CVPR}, 2016.

\bibitem{bao2022beit}
Hangbo Bao, Li Dong, Songhao Piao, and Furu Wei.
\newblock \href{https://arxiv.org/abs/2106.08254}{{BEiT}: {BERT} pre-training
  of image transformers}.
\newblock In {\em ICLR}, 2022.

\bibitem{ARKitScenes}
Gilad Baruch, Zhuoyuan Chen, Afshin Dehghan, Tal Dimry, Yuri Feigin, Peter Fu,
  Thomas Gebauer, Brandon Joffe, Daniel Kurz, Arik Schwartz, and Elad Shulman.
\newblock \href{https://arxiv.org/abs/2111.08897}{ARKitScenes: A diverse
  real-world dataset For 3D indoor scene understanding using mobile RGB-D
  data}.
\newblock In {\em NeurIPS Workshop}, 2021.

\bibitem{brown2020language}
Tom Brown, Benjamin Mann, Nick Ryder, Melanie Subbiah, Jared~D Kaplan, Prafulla
  Dhariwal, Arvind Neelakantan, Pranav Shyam, Girish Sastry, Amanda Askell,
  et~al.
\newblock \href{https://arxiv.org/abs/2005.14165}{Language models are few-shot
  learners}.
\newblock {\em NeurIPS}, 33:1877--1901, 2020.

\bibitem{shapenet2015}
Angel~X. Chang, Thomas Funkhouser, Leonidas Guibas, Pat Hanrahan, Qixing Huang,
  Zimo Li, Silvio Savarese, Manolis Savva, Shuran Song, Hao Su, Jianxiong Xiao,
  Li Yi, and Fisher Yu.
\newblock \href{https://arxiv.org/abs/1512.03012}{ShapeNet: An information-rich
  3D model repository}.
\newblock arXiv:1512.03012 [cs.GR], 2015.

\bibitem{chen2022mixformer}
Qiang Chen, Qiman Wu, Jian Wang, Qinghao Hu, Tao Hu, Errui Ding, Jian Cheng,
  and Jingdong Wang.
\newblock
  \href{https://openaccess.thecvf.com/content/CVPR2022/html/Chen_MixFormer_Mixing_Features_Across_Windows_and_Dimensions_CVPR_2022_paper.html}{MixFormer:
  Mixing features across windows and dimensions}.
\newblock In {\em CVPR}, pages 5249--5259, 2022.

\bibitem{chen2022scaling}
Yukang Chen, Jianhui Liu, Xiaojuan Qi, Xiangyu Zhang, Jian Sun, and Jiaya Jia.
\newblock \href{https://arxiv.org/abs/2206.10555}{LargeKernel3D: Scaling up
  kernels in 3D CNNs}, 2023.

\bibitem{choe2021pointmixer}
Jaesung Choe, Chunghyun Park, Francois Rameau, Jaesik Park, and In~So Kweon.
\newblock \href{https://arxiv.org/abs/2111.11187}{PointMixer: MLP-Mixer for
  point cloud understanding}.
\newblock In {\em ECCV}, 2022.

\bibitem{choy20194d}
Christopher Choy, JunYoung Gwak, and Silvio Savarese.
\newblock \href{https://chrischoy.github.io/publication/minkowskinet/}{4D
  spatio-temporal ConvNets: Minkowski convolutional neural networks}.
\newblock In {\em CVPR}, 2019.

\bibitem{chu2021twins}
Xiangxiang Chu, Zhi Tian, Yuqing Wang, Bo Zhang, Haibing Ren, Xiaolin Wei,
  Huaxia Xia, and Chunhua Shen.
\newblock \href{https://arxiv.org/abs/2104.13840}{Twins: Revisiting the design
  of spatial attention in vision transformers}.
\newblock {\em NeurIPS}, 34:9355--9366, 2021.

\bibitem{dai2017scannet}
Angela Dai, Angel~X. Chang, Manolis Savva, Maciej Halber, Thomas Funkhouser,
  and Matthias Nie{\ss}ner.
\newblock \href{http://www.scan-net.org/}{ScanNet: Richly-annotated 3D
  reconstructions of indoor scenes}.
\newblock In {\em CVPR}, 2017.

\bibitem{devlin2018bert}
Jacob Devlin, Ming-Wei Chang, Kenton Lee, and Kristina Toutanova.
\newblock \href{https://arxiv.org/abs/1810.04805}{Bert: Pre-training of deep
  bidirectional transformers for language understanding}.
\newblock In {\em NAACL}, 2019.

\bibitem{dong2022autoencoders}
Runpei Dong, Zekun Qi, Linfeng Zhang, Junbo Zhang, Jianjian Sun, Zheng Ge, Li
  Yi, and Kaisheng Ma.
\newblock \href{https://arxiv.org/pdf/2212.08320.pdf}{Autoencoders as
  cross-modal teachers: Can pretrained 2D image transformers help 3D
  representation learning?}
\newblock In {\em ICLR}, 2023.

\bibitem{dong2022cswin}
Xiaoyi Dong, Jianmin Bao, Dongdong Chen, Weiming Zhang, Nenghai Yu, Lu Yuan,
  Dong Chen, and Baining Guo.
\newblock
  \href{https://openaccess.thecvf.com/content/CVPR2022/html/Dong_CSWin_Transformer_A_General_Vision_Transformer_Backbone_With_Cross-Shaped_Windows_CVPR_2022_paper.html}{CSWin
  transformer: A general vision transformer backbone with cross-shaped
  windows}.
\newblock In {\em CVPR}, pages 12124--12134, 2022.

\bibitem{dosovitskiy2020image}
Alexey Dosovitskiy, Lucas Beyer, Alexander Kolesnikov, Dirk Weissenborn,
  Xiaohua Zhai, Thomas Unterthiner, Mostafa Dehghani, Matthias Minderer, Georg
  Heigold, Sylvain Gelly, et~al.
\newblock \href{https://arxiv.org/pdf/2010.11929.pdf}{An image is worth 16x16
  words: Transformers for image recognition at scale}.
\newblock In {\em ICLR}, 2021.

\bibitem{guo2021pct}
Meng-Hao Guo, Jun-Xiong Cai, Zheng-Ning Liu, Tai-Jiang Mu, Ralph~R Martin, and
  Shi-Min Hu.
\newblock
  \href{https://link.springer.com/article/10.1007/s41095-021-0229-5}{PCT: Point
  cloud transformer}.
\newblock {\em Computational Visual Media}, 7(2):187--199, 2021.

\bibitem{guo2022attention}
Meng-Hao Guo, Tian-Xing Xu, Jiang-Jiang Liu, Zheng-Ning Liu, Peng-Tao Jiang,
  Tai-Jiang Mu, Song-Hai Zhang, Ralph~R Martin, Ming-Ming Cheng, and Shi-Min
  Hu.
\newblock
  \href{https://link.springer.com/article/10.1007/s41095-022-0271-y}{Attention
  mechanisms in computer vision: A survey}.
\newblock {\em Computational Visual Media}, pages 1--38, 2022.

\bibitem{gwak2020generative}
JunYoung Gwak, Christopher Choy, and Silvio Savarese.
\newblock \href{https://arxiv.org/abs/2006.12356}{Generative sparse detection
  networks for 3D single-shot object detection}.
\newblock In {\em ECCV}, pages 297--313, 2020.

\bibitem{gwak2020gsdn}
JunYoung Gwak, Christopher~B Choy, and Silvio Savarese.
\newblock \href{https://jgwak.com/publications/gsdn/}{Generative sparse
  detection networks for 3D single-shot object detection}.
\newblock In {\em ECCV}, 2020.

\bibitem{han2022survey}
Kai Han, Yunhe Wang, Hanting Chen, Xinghao Chen, Jianyuan Guo, Zhenhua Liu,
  Yehui Tang, An Xiao, Chunjing Xu, Yixing Xu, et~al.
\newblock \href{https://arxiv.org/abs/2012.12556}{A survey on vision
  transformer}.
\newblock {\em IEEE Trans. Pattern Anal. Mach. Intell.}, 2022.

\bibitem{han2021demystifying}
Qi Han, Zejia Fan, Qi Dai, Lei Sun, Ming-Ming Cheng, Jiaying Liu, and Jingdong
  Wang.
\newblock \href{https://arxiv.org/abs/2106.04263}{Demystifying local vision
  transformer: Sparse connectivity, weight sharing, and dynamic weight}.
\newblock In {\em ICLR}, 2022.

\bibitem{HeCXLDG22}
Kaiming He, Xinlei Chen, Saining Xie, Yanghao Li, Piotr Doll{\'{a}}r, and
  Ross~B. Girshick.
\newblock \href{https://arxiv.org/abs/2111.06377}{Masked autoencoders are
  scalable vision learners}.
\newblock In {\em CVPR}, pages 15979--15988, 2022.

\bibitem{hou2020efficient}
Ji Hou, Benjamin Graham, Matthias Nie{\ss}ner, and Saining Xie.
\newblock \href{https://arxiv.org/abs/2012.09165}{Exploring data-efficient 3D
  scene understanding with contrastive scene contexts}.
\newblock In {\em CVPR}, 2021.

\bibitem{huang2022clip2point}
Tianyu Huang, Bowen Dong, Yunhan Yang, Xiaoshui Huang, Rynson~WH Lau, Wanli
  Ouyang, and Wangmeng Zuo.
\newblock \href{https://arxiv.org/abs/2210.01055}{Clip2Point: Transfer clip to
  point cloud classification with image-depth pre-training}.
\newblock arXiv:2210.01055, 2022.

\bibitem{huang2022frozen}
Xiaoshui Huang, Sheng Li, Wentao Qu, Tong He, Yifan Zuo, and Wanli Ouyang.
\newblock \href{https://arxiv.org/abs/2212.04098}{Frozen CLIP model is
  efficient point cloud backbone}.
\newblock arXiv:2212.04098, 2022.

\bibitem{khan2021transformers}
Salman Khan, Muzammal Naseer, Munawar Hayat, Syed~Waqas Zamir, Fahad~Shahbaz
  Khan, and Mubarak Shah.
\newblock \href{https://arxiv.org/abs/2101.01169}{Transformers in vision: A
  survey}.
\newblock {\em ACM Computing Surveys (CSUR)}, 2021.

\bibitem{lahoud20223d}
Jean Lahoud, Jiale Cao, Fahad~Shahbaz Khan, Hisham Cholakkal, Rao~Muhammad
  Anwer, Salman Khan, and Ming-Hsuan Yang.
\newblock \href{https://arxiv.org/abs/2208.04309}{3D vision with transformers:
  A survey}, 2022.

\bibitem{lai2022stratified}
Xin Lai, Jianhui Liu, Li Jiang, Liwei Wang, Hengshuang Zhao, Shu Liu, Xiaojuan
  Qi, and Jiaya Jia.
\newblock \href{https://arxiv.org/abs/2203.14508}{Stratified transformer for 3D
  point cloud segmentation}.
\newblock In {\em CVPR}, pages 8500--8509, 2022.

\bibitem{li2022sepvit}
Wei Li, Xing Wang, Xin Xia, Jie Wu, Xuefeng Xiao, Min Zheng, and Shiping Wen.
\newblock \href{https://arxiv.org/abs/2203.15380}{SepViT: Separable vision
  transformer}.
\newblock arXiv:2203.15380[cs.CV], 2022.

\bibitem{lin2022meta}
Haojia Lin, Xiawu Zheng, Lijiang Li, Fei Chao, Shanshan Wang, Yan Wang,
  Yonghong Tian, and Rongrong Ji.
\newblock \href{https://arxiv.org/pdf/2211.14462.pdf}{Meta architecture for
  point cloud analysis}.
\newblock In {\em CVPR}, 2023.

\bibitem{Liu2022maskdis}
Haotian Liu, Mu Cai, and Yong~Jae Lee.
\newblock \href{https://arxiv.org/abs/2203.11183 }{Masked discrimination for
  self-supervised learning on point clouds}.
\newblock In {\em ECCV}, 2022.

\bibitem{liu2022swin}
Ze Liu, Han Hu, Yutong Lin, Zhuliang Yao, Zhenda Xie, Yixuan Wei, Jia Ning, Yue
  Cao, Zheng Zhang, Li Dong, et~al.
\newblock
  \href{https://openaccess.thecvf.com/content/CVPR2022/html/Liu_Swin_Transformer_V2_Scaling_Up_Capacity_and_Resolution_CVPR_2022_paper.html}{Swin
  transformer V2: Scaling up capacity and resolution}.
\newblock In {\em CVPR}, pages 12009--12019, 2022.

\bibitem{liu2021swin}
Ze Liu, Yutong Lin, Yue Cao, Han Hu, Yixuan Wei, Zheng Zhang, Stephen Lin, and
  Baining Guo.
\newblock \href{https://arxiv.org/abs/2103.14030}{Swin transformer:
  Hierarchical vision transformer using shifted windows}.
\newblock In {\em ICCV}, pages 10012--10022, 2021.

\bibitem{nekrasov2021mix3d}
Alexey Nekrasov, Jonas Schult, Or Litany, Bastian Leibe, and Francis Engelmann.
\newblock \href{https://arxiv.org/abs/2110.02210}{Mix3D: Out-of-context data
  augmentation for 3D scenes}.
\newblock In {\em 3DV}, pages 116--125, 2021.

\bibitem{Pang2022MaskedAF}
Yatian Pang, Wenxiao Wang, Francis E.~H. Tay, W. Liu, Yonghong Tian, and
  Liuliang Yuan.
\newblock \href{https://arxiv.org/abs/2203.06604}{Masked autoencoders for point
  cloud self-supervised learning}.
\newblock In {\em ECCV}, 2022.

\bibitem{park2022fast}
Chunghyun Park, Yoonwoo Jeong, Minsu Cho, and Jaesik Park.
\newblock \href{http://cvlab.postech.ac.kr/research/FPT/}{Fast point
  transformer}.
\newblock In {\em CVPR}, pages 16949--16958, 2022.

\bibitem{qian2022pointnext}
Guocheng Qian, Yuchen Li, Houwen Peng, Jinjie Mai, Hasan Abed Al~Kader Hammoud,
  Mohamed Elhoseiny, and Bernard Ghanem.
\newblock \href{https://arxiv.org/abs/2206.04670}{PointNeXt: Revisiting
  PointNet++ with improved training and scaling strategies}.
\newblock In {\em NeurIPS}, 2022.

\bibitem{ran2022surface}
Haoxi Ran, Jun Liu, and Chengjie Wang.
\newblock \href{https://arxiv.org/abs/2205.05740}{Surface representation for
  point clouds}.
\newblock In {\em CVPR}, pages 18942--18952, 2022.

\bibitem{rao2021randomrooms}
Yongming Rao, Benlin Liu, Yi Wei, Jiwen Lu, Cho-Jui Hsieh, and Jie Zhou.
\newblock \href{https://arxiv.org/abs/2108.07794}{RandomRooms: Unsupervised
  pre-training from synthetic shapes and randomized layouts for 3D object
  detection}.
\newblock In {\em ICCV}, pages 3283--3292, 2021.

\bibitem{rukhovich2021fcaf3d}
Danila Rukhovich, Anna Vorontsova, and Anton Konushin.
\newblock \href{https://arxiv.org/abs/2112.00322}{FCAF3D: Fully convolutional
  anchor-free 3D object detection}.
\newblock In {\em ECCV}, 2021.

\bibitem{shaw2018self}
Peter Shaw, Jakob Uszkoreit, and Ashish Vaswani.
\newblock \href{https://arxiv.org/abs/1803.02155}{Self-attention with relative
  position representations}.
\newblock In {\em NAACL}, 2018.

\bibitem{Thomas2019}
Hugues Thomas, Charles~R. Qi, Jean-Emmanuel Deschaud, Beatriz Marcotegui,
  Fran\c{c}ois Goulette, and Leonidas~J. Guibas.
\newblock \href{https://arxiv.org/abs/1904.08889}{{KPConv}: Flexible and
  deformable convolution for point clouds}.
\newblock In {\em ICCV}, 2019.

\bibitem{tu2022maxvit}
Zhengzhong Tu, Hossein Talebi, Han Zhang, Feng Yang, Peyman Milanfar, Alan
  Bovik, and Yinxiao Li.
\newblock \href{https://arxiv.org/abs/2204.01697}{MaxViT: Multi-axis vision
  transformer}.
\newblock In {\em ECCV}, 2022.

\bibitem{vu2022softgroup}
Thang Vu, Kookhoi Kim, Tung~M Luu, Thanh Nguyen, and Chang~D Yoo.
\newblock \href{https://arxiv.org/pdf/2203.01509.pdf}{SoftGroup for 3D instance
  segmentation on point clouds}.
\newblock In {\em CVPR}, pages 2708--2717, 2022.

\bibitem{wang2022cagroupd}
Haiyang Wang, Lihe Ding, Shaocong Dong, Shaoshuai Shi, Aoxue Li, Jianan Li,
  Zhenguo Li, and Liwei Wang.
\newblock \href{https://arxiv.org/abs/2210.04264}{CAGroup3D: Class-aware
  grouping for 3D object detection on point clouds}.
\newblock In {\em NeurIPS}, 2022.

\bibitem{octformer}
Peng-Shuai Wang.
\newblock \href{https://arxiv.org/abs/2305.03045}{OctFormer: Octree-based
  transformers for 3D point clouds }.
\newblock {\em ACM Trans. Graph.}, 2023.

\bibitem{Wang2017}
Peng-Shuai Wang, Yang Liu, Yu-Xiao Guo, Chun-Yu Sun, and Xin Tong.
\newblock \href{http://wang-ps.github.io/O-CNN_files/CNN3D.pdf}{O-CNN:
  Octree-based convolutional neural networks for 3D shape analysis}.
\newblock {\em ACM Trans. Graph.}, 36(4):72:1--72:11, 2017.

\bibitem{Wang2020a}
Peng-Shuai Wang, Yu-Qi Yang, Qian-Fang Zou, Zhirong Wu, Yang Liu, and Xin Tong.
\newblock \href{https://arxiv.org/abs/2008.01068}{Unsupervised 3D learning for
  shape analysis via multiresolution instance discrimination}.
\newblock In {\em AAAI}, 2020.

\bibitem{wang2022window}
Qi Wang, Sheng Shi, Jiahui Li, Wuming Jiang, and Xiangde Zhang.
\newblock \href{https://arxiv.org/abs/2212.02287}{Window normalization:
  Enhancing point cloud understanding by unifying inconsistent point
  densities}.
\newblock arXiv:2212.02287, 2022.

\bibitem{wang2021crossformer}
Wenxiao Wang, Lu Yao, Long Chen, Binbin Lin, Deng Cai, Xiaofei He, and Wei Liu.
\newblock \href{https://arxiv.org/abs/2108.00154}{Crossformer: A versatile
  vision transformer hinging on cross-scale attention}.
\newblock In {\em ICLR}, 2022.

\bibitem{wang2022p2p}
Ziyi Wang, Xumin Yu, Yongming Rao, Jie Zhou, and Jiwen Lu.
\newblock \href{https://arxiv.org/abs/2208.02812}{P2P: Tuning pre-trained image
  models for point cloud analysis with point-to-pixel prompting}.
\newblock In {\em NeurIPS}, 2022.

\bibitem{wu2021rethinking}
Kan Wu, Houwen Peng, Minghao Chen, Jianlong Fu, and Hongyang Chao.
\newblock \href{https://arxiv.org/abs/2107.14222}{Rethinking and improving
  relative position encoding for vision transformer}.
\newblock In {\em ICCV}, pages 10033--10041, 2021.

\bibitem{wu2022pale}
Sitong Wu, Tianyi Wu, Haoru Tan, and Guodong Guo.
\newblock \href{https://ojs.aaai.org/index.php/AAAI/article/view/20176}{Pale
  transformer: A general vision transformer backbone with pale-shaped
  attention}.
\newblock In {\em AAAI}, pages 2731--2739, 2022.

\bibitem{wu2022pointconvformer}
Wenxuan Wu, Qi Shan, and Li Fuxin.
\newblock \href{https://arxiv.org/abs/2208.02879}{PointConvFormer: Revenge of
  the point-based convolution}.
\newblock In {\em CVPR}, 2023.

\bibitem{wu2022point}
Xiaoyang Wu, Yixing Lao, Li Jiang, Xihui Liu, and Hengshuang Zhao.
\newblock \href{https://arxiv.org/abs/2210.05666}{Point Transformer V2: Grouped
  vector attention and partition-based pooling}.
\newblock In {\em NeurIPS}, 2022.

\bibitem{wu2023masked}
Xiaoyang Wu, Xin Wen, Xihui Liu, and Hengshuang Zhao.
\newblock \href{https://arxiv.org/abs/2303.14191}{Masked scene contrast: A
  scalable framework for unsupervised 3D representation learning}.
\newblock In {\em CVPR}, pages 9415--9424, 2023.

\bibitem{Xie2020}
Saining Xie, Jiatao Gu, Demi Guo, Charles~R Qi, Leonidas~J Guibas, and Or
  Litany.
\newblock \href{https://arxiv.org/abs/2007.10985}{PointContrast: Unsupervised
  pre-training for 3D point cloud understanding}.
\newblock {\em ECCV}, 2020.

\bibitem{yang2021focal}
Jianwei Yang, Chunyuan Li, Pengchuan Zhang, Xiyang Dai, Bin Xiao, Lu Yuan, and
  Jianfeng Gao.
\newblock \href{https://arxiv.org/abs/2107.00641}{Focal self-attention for
  local-global interactions in vision transformers}.
\newblock {\em NeurIPS}, 34:30008--30022, 2021.

\bibitem{Yu_2022_CVPR}
Xumin Yu, Lulu Tang, Yongming Rao, Tiejun Huang, Jie Zhou, and Jiwen Lu.
\newblock \href{https://arxiv.org/abs/2111.14819}{{Point-BERT}: Pre-training 3D
  point cloud transformers with masked point modeling}.
\newblock In {\em CVPR}, pages 19313--19322, June 2022.

\bibitem{yuan2021hrformer}
Yuhui Yuan, Rao Fu, Lang Huang, Weihong Lin, Chao Zhang, Xilin Chen, and
  Jingdong Wang.
\newblock
  \href{https://proceedings.neurips.cc/paper/2021/hash/3bbfdde8842a5c44a0323518eec97cbe-Abstract.html}{HRFormer:
  High-resolution vision transformer for dense prediction}.
\newblock {\em NeurIPS}, 34:7281--7293, 2021.

\bibitem{zhang2022vsa}
Qiming Zhang, Yufei Xu, Jing Zhang, and Dacheng Tao.
\newblock \href{https://arxiv.org/abs/2204.08446}{VSA: Learning varied-size
  window attention in vision transformers}.
\newblock In {\em ECCV}, 2022.

\bibitem{zhang2022point}
Renrui Zhang, Ziyu Guo, Peng Gao, Rongyao Fang, Bin Zhao, Dong Wang, Yu Qiao,
  and Hongsheng Li.
\newblock \href{https://arxiv.org/abs/2205.14401}{Point-M2AE: Multi-scale
  masked autoencoders for hierarchical point cloud pre-training}.
\newblock In {\em NeurIPS}, 2022.

\bibitem{zhang2021self}
Zaiwei Zhang, Rohit Girdhar, Armand Joulin, and Ishan Misra.
\newblock \href{https://arxiv.org/abs/2101.02691}{Self-supervised pretraining
  of 3D features on any point-cloud}.
\newblock In {\em ICCV}, pages 10252--10263, 2021.

\bibitem{zhao2021point}
Hengshuang Zhao, Li Jiang, Jiaya Jia, Philip~HS Torr, and Vladlen Koltun.
\newblock
  \href{https://openaccess.thecvf.com/content/ICCV2021/html/Zhao_Point_Transformer_ICCV_2021_paper.html}{Point
  transformer}.
\newblock In {\em ICCV}, pages 16259--16268, 2021.

\bibitem{zheng2020structured3d}
Jia Zheng, Junfei Zhang, Jing Li, Rui Tang, Shenghua Gao, and Zihan Zhou.
\newblock \href{https://structured3d-dataset.org/}{Structured3D: A large
  photo-realistic dataset for structured 3D modeling}.
\newblock In {\em ECCV}, pages 519--535, 2020.

\bibitem{zou2021manhattan}
Chuhang Zou, Jheng-Wei Su, Chi-Han Peng, Alex Colburn, Qi Shan, Peter Wonka,
  Hung-Kuo Chu, and Derek Hoiem.
\newblock \href{https://arxiv.org/abs/1910.04099}{Manhattan room layout
  reconstruction from a single 360 image: A Comparative study of
  state-of-the-art methods}.
\newblock {\em International Journal of Computer Vision}, 129:1410--1431, 2021.

\end{thebibliography}
}

\appendix

\subsection*{Appendix: Efficient self-attention implementation} \label{app:sec:implementation}
Apart from our memory-efficient self-attention approach, we have also enhanced our self-attention computation in the following ways.

\myparagraph{Kernel scheduling}
Designing an optimal scheduling strategy for CUDA kernels is essential to make the most of the memory bandwidth and computational capabilities of modern GPUs. Stratified Transformer~\cite{lai2022stratified} builds CUDA blocks whose number is equal to the number of points in the window and creates GPU threads whose number is the maximum number of points in the window. However, the varying number of points in each window makes it difficult for the CUDA compiler to optimize the execution speed. To tackle this issue, our kernel is designed to calculate the channel-wise contribution to the weight coefficients $c_{ij,h,d}$ ($i,j$ are the indices of queries and keys, $h$ is the index of the head, $d$ is the index of the channel) per thread and binds blocks to a fixed number of threads ($256$ in our implementation). We use a local synchronized operation (\textit{\_\_shfl\_down\_sync}) to perform \textit{reduce-sum} to obtain the coefficient weight $c_{ij,h}$. Thanks to the compacted memory access between neighboring threads and fixed and balanced operations per thread, our implementation results in higher memory bandwidth and efficient computation.

\myparagraph{Half-precision support} We enable half-precision for all CUDA kernels that we have implemented. Additionally, we have reorganized the memory layout of look-up tables from dimension-last to channel-last. This allows a single thread to process two consecutive elements with a single $32$-bit memory access for the query, key, and look-up tables, respectively.

\myparagraph{Computation speedup} We combine the calculation of coefficient weights and cRPE/cRSE into a single kernel, thus decreasing the memory access of queries and keys and speeding up GPU performance.

\myparagraph{Reduction of atomic operations} We employ atomic operations to compute the updated features and gradients of the lookup tables. However, half-precision atomic operators can be inefficient due to double writing/reading conflicts. To address this issue, we use shared memory to combine two consecutive $16$-bit atomic operations into a single $32$-bit atomic operation, thus reducing the number of atomic operations and speeding up GPU execution. 

\end{document}